\documentclass{article}

\PassOptionsToPackage{round,sort,comma}{natbib}
\usepackage[final]{neurips_2021}

\usepackage[utf8]{inputenc} 
\usepackage[T1]{fontenc}    
\usepackage{url}            

\usepackage{algorithm}
\usepackage{algorithmic}

\usepackage[export]{adjustbox}
\usepackage{subcaption}
\usepackage[percent]{overpic}
\usepackage{wrapfig}
\usepackage{graphicx}
\usepackage{svg}

\usepackage{pifont}
\usepackage{float}

\usepackage{amsmath,amsfonts,amsthm,amssymb, nccmath}
\usepackage{nicefrac}       
\allowdisplaybreaks 
\usepackage{bbm}
\usepackage{framed}
\usepackage{enumerate}
\usepackage{mathtools}
\usepackage{scalefnt}
\usepackage{changepage}

\usepackage{comment}
\usepackage{setspace}
\usepackage{verbatim}
\usepackage{xcolor}
\usepackage{scalefnt}
\usepackage{xr}

\usepackage[shortlabels]{enumitem}

\usepackage{placeins}
\usepackage{color}
\usepackage{footmisc}
\usepackage{xspace}

\usepackage{thmtools}
\usepackage{thm-restate}

\usepackage{multirow}
\usepackage{booktabs}
\usepackage{array}
\usepackage{tabularx}

\usepackage{microtype}

\usepackage[colorlinks=true, citecolor=blue, linkcolor=black]{hyperref}

\usepackage{tikz}
\usetikzlibrary{calc}
\newcommand{\paragraphsmall}[1]{\textbf{{#1} $\;$}}

\makeatletter
\newcommand{\settitle}{\@maketitle}
\makeatother

\DeclarePairedDelimiterX{\dotp}[2]{\langle}{\rangle}{#1, #2}

\makeatletter
\newcommand*\bigcdot{\mathpalette\bigcdot@{.5}}
\newcommand*\bigcdot@[2]{\mathbin{\vcenter{\hbox{\scalebox{#2}{$\m@th#1\bullet$}}}}}
\makeatother

\newcommand{\be}{\begin{equation}}
\newcommand{\ee}{\end{equation}}

\newcommand{\norm}[1]{\left\lVert#1\right\rVert}

\newcommand{\abs}[1]        {| #1 |}

\renewcommand{\epsilon}{\varepsilon}

\newcommand{\na}{N/A}

\renewcommand{\paragraph}{\paragraphsmall}

\title{Sparse Flows: Pruning Continuous-depth Models}

\newcommand{\setauthors}[1][$^{*}$]{
\author{%
    Lucas Liebenwein{#1} \\
    MIT CSAIL \\
    \texttt{lucas@csail.mit.edu} \\
    \And
    Ramin Hasani$^{*}$ \\ 
    MIT CSAIL \\
    \texttt{rhasani@mit.edu} \\
    \And
    Alexander Amini \\
    MIT CSAIL \\
    \texttt{amini@mit.edu} \\
    \And
    Daniela Rus \\
    MIT CSAIL \\
    \texttt{rus@csail.mit.edu} \\
}
}

\setauthors[\thanks{denotes authors with equal contributions. Code: \url{https://github.com/lucaslie/torchprune}}]

\begin{document}

\maketitle

\begin{abstract}
Continuous deep learning architectures enable learning of flexible probabilistic models for predictive modeling as neural ordinary differential equations (ODEs), and for generative modeling as continuous normalizing flows. In this work, we design a framework to decipher the internal dynamics of these continuous depth models by pruning their network architectures. Our empirical results suggest that pruning improves generalization for neural ODEs in generative modeling. We empirically show that the improvement is because pruning helps avoid mode-collapse and flatten the loss surface. Moreover, pruning finds efficient neural ODE representations with up to 98\% less parameters compared to the original network, without loss of accuracy. We hope our results will invigorate further research into the performance-size trade-offs of modern continuous-depth models.
\end{abstract}

\begin{wrapfigure}[14]{r}{0.5\textwidth}
\vspace{-5mm}
	\centering
	\includegraphics[width=0.5\textwidth]{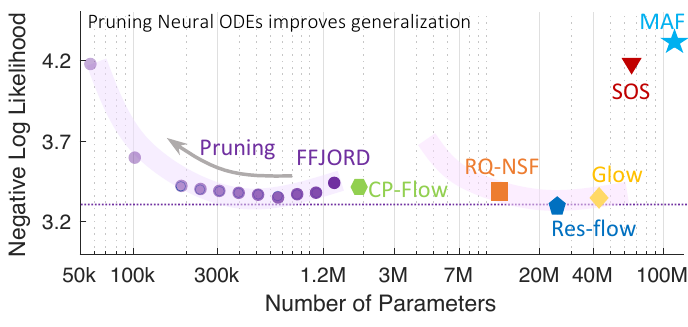}
	\vspace{-5mm}
	\caption{Pruning Neural ODEs improves their generalization with at least 1 order of magnitude less parameters. CIFAR-10 density estimation. Values and methods are described in Table \ref{tab:image_results}.}
	\label{fig:intro}
\end{wrapfigure}
\section{Introduction}
\label{sec:introduction}
The continuous analog of normalizing flows (CNFs) \citep{chen2018neural} efficiently \citep{grathwohl2019ffjord} maps a latent space to data by ordinary differential equations (ODEs), relaxing the strong constraints over \emph{discrete} normalizing flows \citep{rezende2015variational,dinh2016density,papamakarios2017masked,kingma2018glow,durkan2019neural,huang2020convex}. CNFs enable learning flexible models by unconstrained neural networks. While recent works investigated ways to improve CNFs' efficiency \citep{grathwohl2019ffjord,finlay2020train,li2020scalable}, regularize the flows \citep{yang2020potential,onken2020ot}, or solving their shortcomings such as crossing trajectories \citep{dupont2019augmented,massaroli2020dissecting}, less is understood about their inner dynamics during and post training.

In this paper, we set out to use standard pruning algorithms to investigate generalization properties of sparse neural ODEs and continuous normalizing flows. In particular, we investigate how the inner dynamics and the modeling performance of a continuous flow varies if we methodologically prune its neural network architecture. Reducing unnecessary weights of a neural network (pruning) \citep{lecun1990optimal,hassibi1993second,han2015learning,li2016pruning} without loss of accuracy results in smaller network size \citep{hinton2015distilling, liebenwein2021lost, liebenwein2021compressing}, computational efficiency \citep{yang2017designing,molchanov2019pruning,luo2017thinet}, faster inference \citep{frankle2018lottery}, and enhanced interpretability \citep{liebenwein2020provable,lechner2020neural, baykal2018datadependent, sipp2019}. Here, our main objective is to better understand CNFs' dynamics in density estimation tasks as we increase network sparsity and to show that pruning can improve generalization in neural ODEs.

\begin{wrapfigure}[14]{r}{0.4\textwidth}
\vspace{-1mm}
	\centering
	\includegraphics[width=0.4\textwidth]{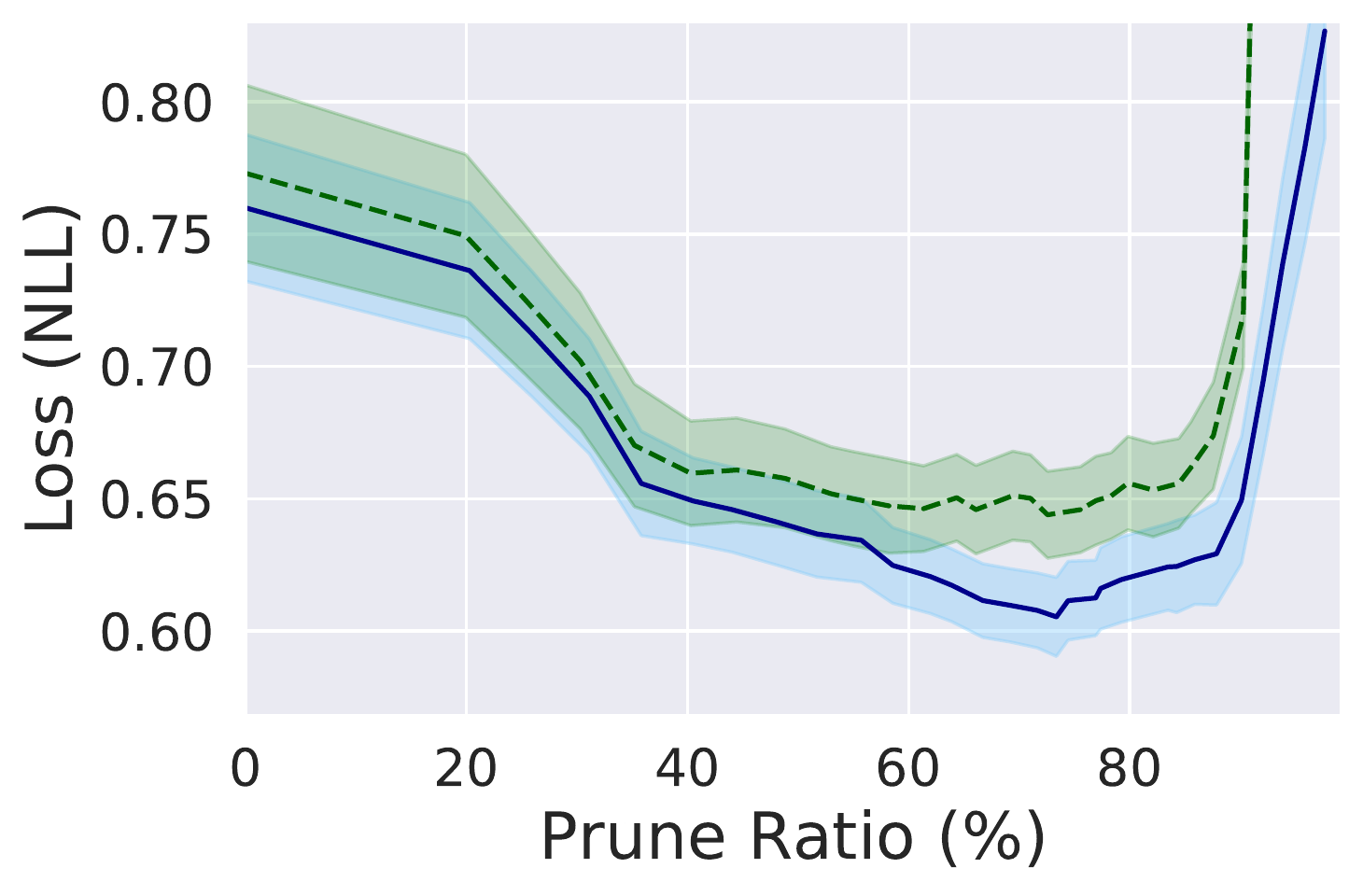}
	\vspace{-5mm}
	\caption{Pruning enhances generalization of continuous-depth models. Structured pruning (green), unstructured pruning (blue). More details in Section \ref{sec:experiments}.}
	\label{fig:intro-u-curve}
\end{wrapfigure}
\paragraph{Pruning improves generalization in neural ODEs.} Our results consistently suggest that a certain ratio of pruning of fully connected neural ODEs leads to lower empirical risk in density estimation tasks, thus obtaining better generalization. We validate this observation on a large series of experiments with increasing dimensionality. See an example here in Figure \ref{fig:intro-u-curve}.

\begin{wrapfigure}[12]{r}{0.4\textwidth}
\vspace{-6mm}
	\centering
	\includegraphics[width=0.3\textwidth]{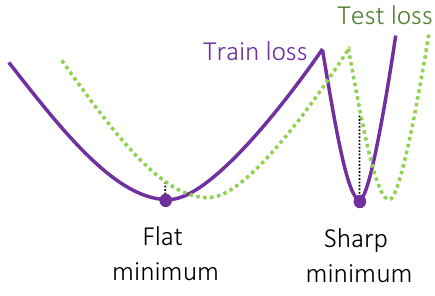}
	\vspace{-2mm}
	\caption{Flat minima result in better generalization compared to sharp minima. Pruning neural ODEs flattens the loss around local minima. Figure is reproduced from \citet{keskar2017large}.}
	\label{fig:flas-vs-sharp-minimum}
\end{wrapfigure}
\textbf{Pruning flattens the loss surface of neural ODEs.} Additionally, we conduct a Hessian-based empirical investigation on the objective function of the flows-under-test in density estimation tasks to better understand why pruning results in better generalization. We find that for Neural ODEs, pruning decreases the value of the Hessian's eigenvalues, and as a result, flattens the loss which leads to better generalization, c.f., \cite{keskar2017large} (Figure \ref{fig:flas-vs-sharp-minimum}).

\textbf{Pruning helps avoiding mode-collapse in generative modeling.} In a series of multi-modal density estimation tasks, we observe that densely connected CNFs often get stuck in a sharp local minimum (See Figure \ref{fig:flas-vs-sharp-minimum}) and as a result, cannot properly distinguish different modes of data. This phenomena is known as mode-collapse. Once we sparsify the flows, the quality of the density estimation task increases significantly and consequently mode-collapse does not occur.

\textbf{Pruning finds minimal and efficient neural ODE representations.} Our framework finds highly optimized and efficient neural ODE architectures via pruning. In many instances, we can reduce the parameter count by 70-98\% (6x-50x compression rate). Notably, one cannot directly train such sparse and efficient continuous-depth models from scratch.

\section{Background}
\label{sec:preliminaries}
In this section, we describe the necessary background to construct our framework. 
We show how to perform generative modeling by continuous depth models using the change of variables formula.

\paragraph{Generative modeling via change of variables.} 
The change of variables formula uses an invertible mapping $f: \mathbb{R}^D \rightarrow \mathbb{R}^D$, to wrap a normalized base distribution $p_z(\textbf{z})$, to specify a more complex distribution. In particular, given $z \sim p_z(\textbf{z})$, a random variable, the log density for function $f(\textbf{z})=\textbf{x}$ can be computed by \citet{dinh2015nice}:
\begin{equation}
\label{eq:change_of_variables}
    \log p_x(\textbf{x}) = \log p_z(\textbf{z}) - \log \det \left| \frac{\partial f(\textbf{z})}{\partial \textbf{z}}\right|,
\end{equation}
where $\frac{\partial f(\textbf{z})}{\partial \textbf{z}}$ is the Jacobian of $f$. While theoretically Eq. \ref{eq:change_of_variables} demonstrates a simple way to finding the log density, from a practical standpoint computation of the Jacobian determinant has a time complexity of $\mathcal{O}(D^3)$. Restricting network architectures can make its computation more tractable. Examples include designing normalizing flows \citep{rezende2015variational,papamakarios2017masked,berg2018sylvester}, autoregressive transformations \citep{kingma2016improving,oliva2018transformation,muller2019neural,wehenkel2019unconstrained,durkan2019neural,jaini2019sum}, partitioned transformations \citep{dinh2016density,kingma2018glow}, universal flows \citep{kong2020expressive,teshima2020coupling}, and the use of optimal transport theorem \citep{huang2020convex}.

Alternative to these discrete transformation algorithms, one can construct a generative model similar to (\ref{eq:change_of_variables}), and declare $f$ by a continuous-time dynamics \citep{chen2018neural,grathwohl2019ffjord,lechner2020gershgorin}. Given a sample from the base distribution, one can parametrize an ordinary differential equations (ODEs) by a function $f(\textbf{z}(t),t,\theta)$, and solve the ODE to obtain the observable data. When $f$ is a neural network, the system is called a neural ODE \citep{chen2018neural}. 

\paragraph{Neural ODEs.} 
More formally, a neural ODE is defined by finding the solution to the initial value problem (IVP): $\frac{\partial \textbf{z}(t)}{\partial t} = f(\textbf{z}(t),t,\theta)$, $\textbf{z}(t_0) = \textbf{z}_0$, with $\textbf{z}_0 \sim p_{z_0}(\textbf{z}_0)$, to get $\textbf{z}(t_n)$ the desired output observations at a terminal integration step $n$ \citep{chen2018neural}.\footnote{One can design a more expressive representation \citep{hasani2020natural,vorbach2021causal} of continuous-depth models by using the second-order approximation of the neural ODE formulation \citep{Hasani2021liquid}. This representation might give rise to a better neural flows which will be the focus of our continued effort.} 

\paragraph{Continuous Normalizing Flows.} 
If $\textbf{z}(t_n)$ is set to our observable data, given samples from the base distribution $\textbf{z}_0 \sim p_{z_0}(\textbf{z}_0)$, the neural ODE described above forms a continuous normalizing flow (CNF). CNFs modify the change in log density by the left hand-side differential equation and as a result the total change in log-density by the right hand-side equation \citep{chen2018neural,grathwohl2019ffjord}:

\begin{equation}
\label{eq:change_in_log_density}
\frac{\partial \log p(\textbf{z}(t))}{\partial t} = - \text{Tr}\left( \frac{\partial f}{\partial \textbf{z}(t)}\right),~~~ \log p(\textbf{z}(t_n)) = \log p(\textbf{z}(t_0)) - \int_{t_{0}}^{t_1} \text{Tr}\left( \frac{\partial f}{\partial \textbf{z}(t)}\right) dt.
\end{equation}
The system of two differential equations (the neural ODE ($\frac{\partial \textbf{z}(t)}{\partial t} = f(\textbf{z}(t),t,\theta)$) and \eqref{eq:change_in_log_density} can then be solved by automatic differentiation algorithms such as backpropagation through time \citep{rumelhart1986learning,Hasani2021liquid} or the adjoint sensitivity method \citep{pontryagin2018mathematical,chen2018neural}. Computation of $\text{Tr}\left( \frac{\partial f}{\partial \textbf{z}(t)}\right)$ costs $\mathcal{O}(D^2)$. A method called the Free-form Jacobian of Reversible Dynamics (FFJORD) \citep{grathwohl2019ffjord} improved the cost to $\mathcal{O}(D)$ by using the Hutchinson's trace estimator \citep{hutchinson1989stochastic,adams2018estimating}. Thus, the trace of the Jacobian can be estimated by: $\text{Tr} \left( \frac{\partial f}{\partial \textbf{z}(t)} \right) = \mathbb{E} _{ p(\boldsymbol{\epsilon})} \left[ \boldsymbol{\epsilon}^T \frac{\partial f}{\partial \textbf{z}(t)} \boldsymbol{\epsilon} \right]$, where $p(\epsilon)$ is typically set to a Gaussian or Rademacher distribution \citep{grathwohl2019ffjord}. Throughout the paper, we investigate the properties of FFJORD CNFs by pruning their neural network architectures.

\section{Pruning Neural ODEs}
\label{sec:pruning_nodes}

We enable sparsity in neural ODEs and CNFs by removing, i.e., pruning, redundant weights from the underlying neural network architecture during training. Pruning can tremendously improve the parameter efficiency of neural networks across numerous tasks, such as computer vision~\citep{liebenwein2020provable} and natural language processing~\citep{maalouf2021deep}. 

\subsection{A General Framework for Training Sparse Flows}
Our approach to training Sparse Flows is inspired by \emph{iterative learning rate rewinding}, a recently proposed and broadly adopted pruning framework as used by~\citet{renda2020comparing, liebenwein2020provable} among others. 

In short, our pruning framework proceeds by first training an unpruned, i.e., dense network to obtain a warm initialization for pruning. Subsequently, we proceed by iteratively pruning and retraining the network until we either obtain the desired level of sparsity, i.e., prune ratio or when the loss for a pre-specified hold-out dataset (validation loss) starts to deteriorate (early stopping). We note that our framework is readily applicable to any continuous-depth model and not restricted to FFJORD-like models. Moreover, we can account for various types of pruning, i.e., unstructured pruning of weights and structured pruning of neurons or filters. 
An overview of the framework is provided in Algorithm~\ref{alg:prune} and we provide more details below.

\subsection{From Dense to Sparse Flows}
 
\paragraph{\textsc{Train} a dense flow for a warm initialization.}
To initiate the training process, we first train a densely-connected network to obtain a warm initialization~(Line 2 of Algorithm~\ref{alg:prune}). We use Adam with a fixed step learning decay schedule and weight decay in some instances. Based on the warm initialization, we start pruning the network.

\paragraph{\textsc{Prune} for Sparse Flow.}
For the prune step (Line~5 of Algorithm~\ref{alg:prune}) we either consider unstructured or structured pruning, i.e., weight or neuron/filter pruning, respectively. At a fundamental level, unstructured pruning aims at inducing sparsity into the parameters of the flow while structured pruning enables reducing the dimensionality of each flow layer. 

\begin{wraptable}{r}{0.44\textwidth}
\vspace{-2ex}
\centering
\caption{Pruning Methods.}
\label{tab:prune-algs}
\vspace{-1.5ex}
\renewcommand{\arraystretch}{1.05} 
\begin{tabular}{lll}
    \toprule
    & \multirow{2}{*}{\shortstack[l]{Unstructured \\ {\tiny\citep{Han15}}}} 
    & \multirow{2}{*}{\shortstack[l]{Structured \\ {\tiny\citep{li2016pruning}}}} \\
    & \\ \midrule
    \textbf{Target} & Weights & Neurons \\
    \textbf{Score} & $\abs{W_{ij}}$ & $\norm{W_{i :}}_1$ \\
    \textbf{Scope} & Global & Local \\
    \bottomrule
\end{tabular}
\vspace{-2ex}
\end{wraptable}
For unstructured pruning, we use magnitude pruning~\citep{Han15}), where we prune weights across all layers (global) with magnitudes below a pre-defined threshold.
For structured pruning, we use the $\ell_1$-norm of the weights associated with the neuron/filter and prune the structures with lowest norm for constant per-layer prune ratio (local) as proposed by~\citet{li2016pruning}. See Table~\ref{tab:prune-algs} for an overview.

\begin{algorithm}[t!]
\caption{\textsc{SparseFlow}($f$, $\Phi_\text{train}$, $PR$, $e$)}
\label{alg:prune}
\textbf{Input:} 
$f$: neural ODE model with parameter set $\theta$;
$\Phi_\text{train}$: hyper-parameters for training;
$PR$: relative prune ratio; 
$e$: number of training epochs per prune-cycle.
\\
\textbf{Output:} 
$f(\cdot; \hat\theta)$: Sparse Flow;
$m$: sparse connection pattern.
\begin{spacing}{1.1}
\begin{algorithmic}[1]
\STATE $\theta_0 \gets \textsc{RandomInit()}$ \label{lin:randominit} \\
\STATE $\theta \gets \textsc{Train}(\theta_0, \Phi_\text{train}, e)$ \label{lin:train} \COMMENT{Initial training stage with dense neural ODE (``warm start'').} \\
\STATE $m \gets 1^{|\theta_0|}$ \COMMENT{Initialize binary mask indicating neural connection pattern.} \\
\WHILE{validation loss of Sparse Flow decreases} \label{lin:prune-start}
    \STATE $m \gets \textsc{Prune}(m \odot \theta, PR)$ \COMMENT{Prune $PR\%$ of the \emph{remaining} parameters and update mask.} \label{lin:prune} \\
    \STATE $\theta \gets \textsc{Train}(m \odot \theta, \Phi_\text{train}, e)$ \label{lin:retrain} \COMMENT{Restart training with updated connection pattern.}
\ENDWHILE \label{lin:prune-end}
\STATE $\hat \theta \gets m \odot \theta$, ~~~ and ~~~ \textbf{return} $f(\cdot; \hat\theta)$, $m$
\end{algorithmic}
\end{spacing}
\end{algorithm}

\paragraph{\textsc{Train} the Sparse Flow.}
Following the pruning step, we re-initiate the training with the new sparsity pattern and the unpruned weights~(Line~\ref{lin:retrain} of Algorithm~\ref{alg:prune}). Note that we do not change the training hyperparameters between the different stages of training.

\paragraph{\textsc{Iterate} for increased sparsity and performance.}
Naturally, we can iteratively repeat the \textsc{Prune} and \textsc{Train} step to further sparsify the flow (Lines~\ref{lin:prune-start}-\ref{lin:prune-end} of Algorithm~\ref{alg:prune}). Moreover, the resulting sparsity-performance trade-off is affected by the total number of iterations, the relative prune ratio $PR$ per iteration, and the amount of training between \textsc{Prune} steps. We find that a good trade-off is to keep the amount of training constant across all iterations and tune it such that the initial, dense flow is essentially trained to (or close to) convergence. Depending on the difficulty of the task and the available compute resources we can then adapt the per-iteration prune ratio $PR$. Note that the overall relative sparsity after $n$ iterations is given by $(1 - PR)^n$. Detailed hyperparameters and more explanations for each experiment are provided in the supplementary material.

\section{Experiments}
\label{sec:experiments}
We perform a diverse set of experiments demonstrating the effect of pruning on the generalization capability of continuous-depth models. Our experiments include pruning ODE-based flows in density estimation tasks with increasing complexity, as well as pruning neural-ODEs in supervised inference tasks. The density estimation tasks were conducted on flows equipped with Free-form Jacobian of Reversible Dynamics (FFJORD) \citep{grathwohl2019ffjord}, using adaptive ODE solvers \citep{dormand1980family}. We used two code bases (FFJORD from \cite{grathwohl2019ffjord} and TorchDyn \citep{poli2020torchdyn}) over which we implemented our pruning framework.\footnote{All code and data are available online at: \url{https://github.com/lucaslie/torchprune}}

\textbf{Baselines.} In complex density estimation tasks we compare the performance of Sparse Flows to a variety of baseline methods including: FFJORD~\citep{grathwohl2019ffjord}, masked autoencoder density estimation (MADE)~\citep{pmlr-v37-germain15}, Real NVP~\citep{dinh2016density}, masked autoregressive flow (MAF)~\citep{papamakarios2017masked}, Glow \citep{kingma2018glow}, convex potential flows (CP-Flow)~\citep{huang2020convex}, transformation autoregressive networks (TAN)~\citep{pmlr-v80-oliva18a}, neural autoregressive flows (NAF)~\citep{pmlr-v80-huang18d}, and sum-of-squares polynomial flow (SOS)~\citep{jaini2019sum}.

\begin{figure}[t]
\centering
\begin{minipage}[t]{0.52\textwidth}
    \includegraphics[width=\textwidth]{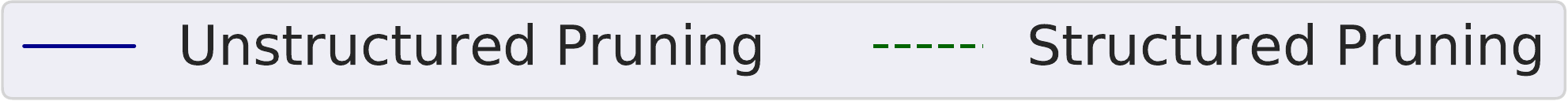}
\end{minipage}
\begin{minipage}[t]{0.33\textwidth}
    \includegraphics[width=\textwidth]{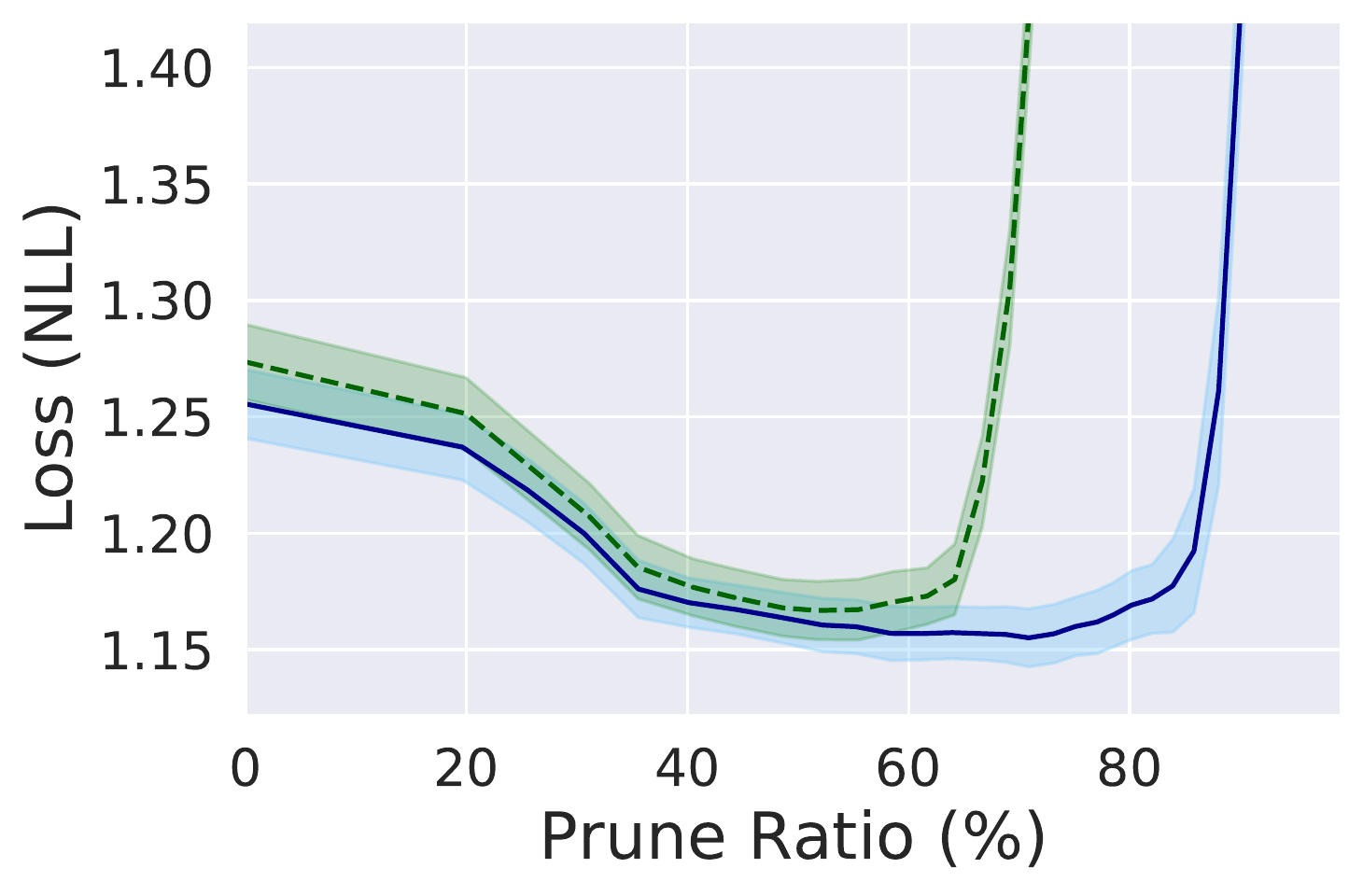}
    \subcaption{Gaussians}
\end{minipage}%
\begin{minipage}[t]{0.33\textwidth}
    \includegraphics[width=\textwidth]{fig/ToyGaussiansSpiral/ffjord_l4_h64_sigmoid_da_ToyGaussiansSpiral_e100/loss/WT_FT.pdf}
    \subcaption{Gaussian Spiral}
\end{minipage}%
\begin{minipage}[t]{0.33\textwidth}
    \includegraphics[width=\textwidth]{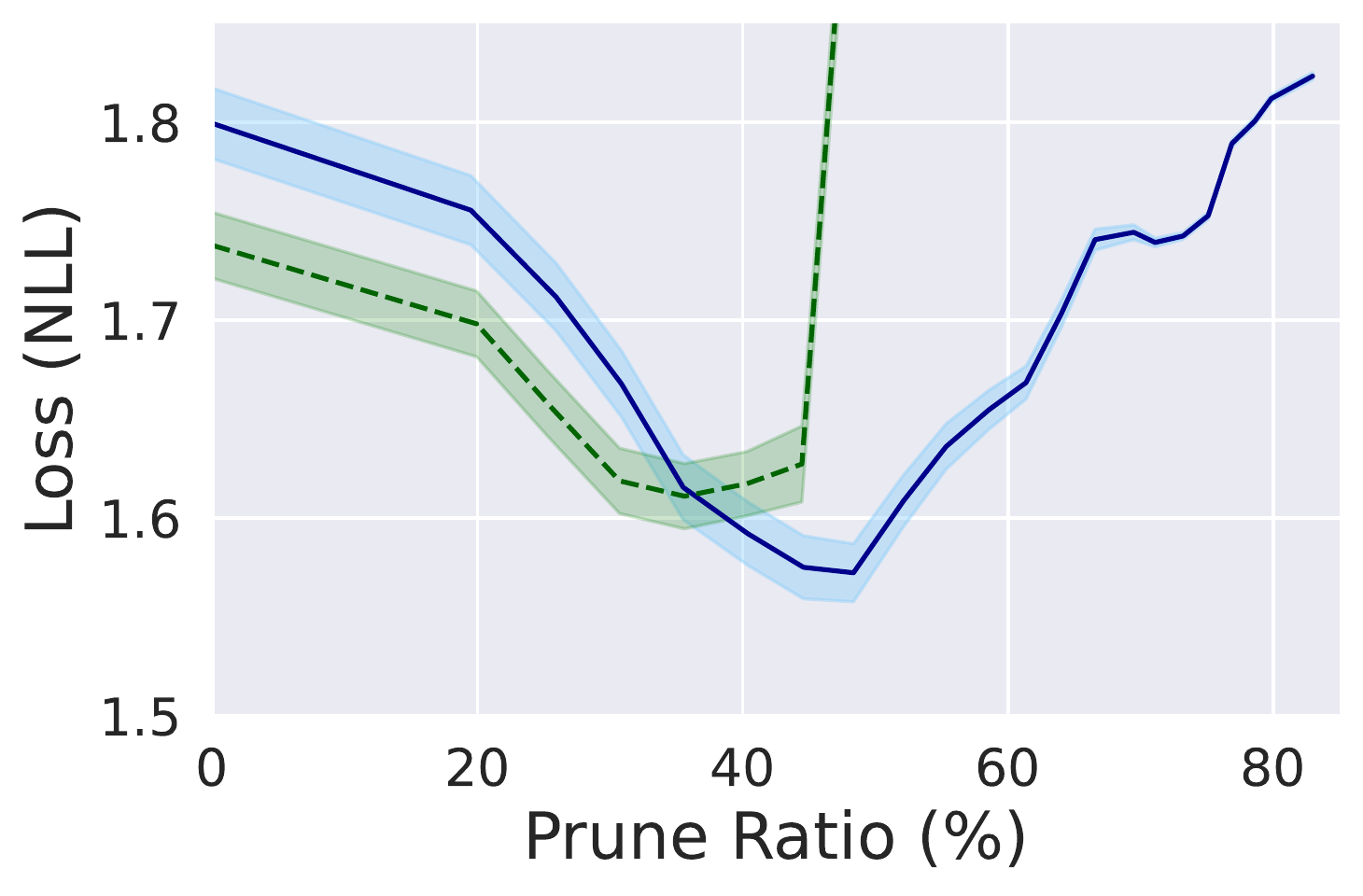}
    \subcaption{Spiral}
\end{minipage}%
\caption{Negative log likelihood of Sparse Flow as function of prune ratio. In general, structured pruning is a more constrained problem as we constrain the type of sparsity. Hence, for almost any pruning experiment we can expect that structured pruning performs worse than unstructured pruning starting at a certain prune ratio. 
}
\label{fig:toy_nll}
\end{figure}

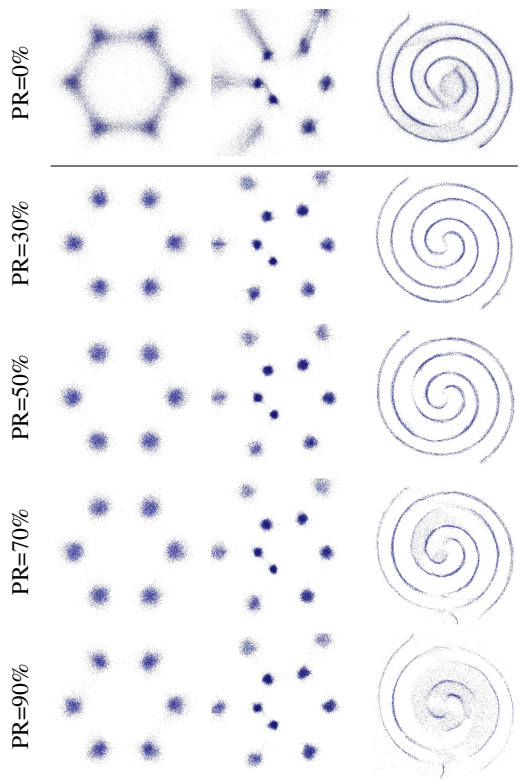
\begin{wrapfigure}[25]{r}{0.52\textwidth}
\vspace{-15mm}
\centering
\begin{tikzpicture}[scale=0.92]

\footnotesize

\newcommand{\tikzfigheight}{55pt}

\coordinate (sepstart) at (0.0,2.1);

\coordinate (textoffset) at (0.2,0);
\coordinate (picoffset) at (0.2,0);

\coordinate (lineoffset) at (0.0, -0.15);

\newcommand{\insertplotrow}[2]{
\node [rotate=90, below left,align=center] (prtext) at ($(sepstart)$) {#1};

\node[right, inner sep=0pt] (plot1) at ($(prtext.south) + (textoffset)$) {
\includegraphics[height=\tikzfigheight]{fig/ToyGaussians/ffjord_l2_h128_sigmoid_da_ToyGaussians_e100/distribution/#2.jpg}
};
\node[right, inner sep=0pt] (plot2) at ($(plot1.east) + (picoffset)$) {
\includegraphics[height=\tikzfigheight]{fig/ToyGaussiansSpiral/ffjord_l4_h64_sigmoid_da_ToyGaussiansSpiral_e100/distribution/#2.jpg}
};
\node[right, inner sep=0pt] (plot3) at ($(plot2.east) + (picoffset)$) {
\includegraphics[height=\tikzfigheight]{fig/ToySpirals2/ffjord_l4_h64_sigmoid_da_ToySpirals2_e100/distribution/#2.jpg}
};

\coordinate (sepstart) at ($(sepstart) - (0,\tikzfigheight) + 2*(lineoffset)$);
}

\newcommand{\drawsepline}{
\draw ($(plot1.south west) + (lineoffset)$) -- ($(plot3.south east) + (lineoffset)$);
}

\insertplotrow{PR=0\%}{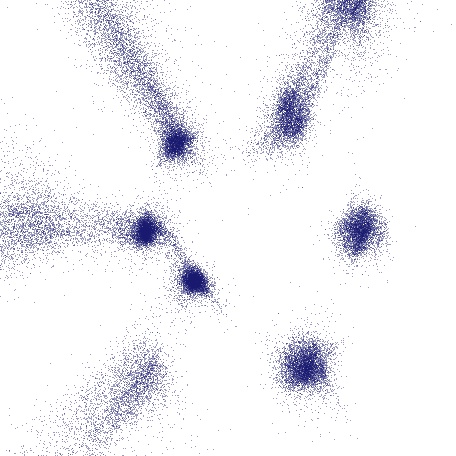};
\drawsepline;
\insertplotrow{PR=30\%}{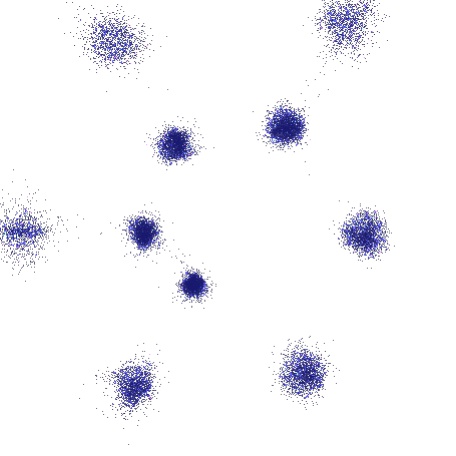};
\insertplotrow{PR=50\%}{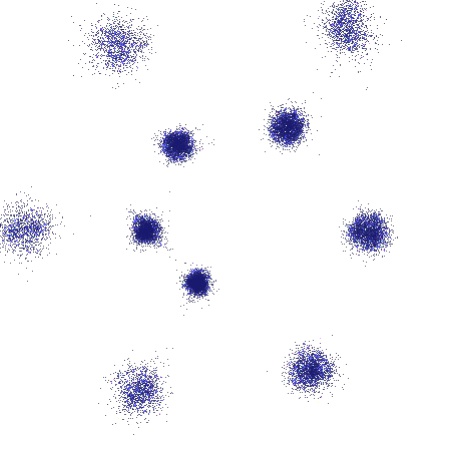};
\insertplotrow{PR=70\%}{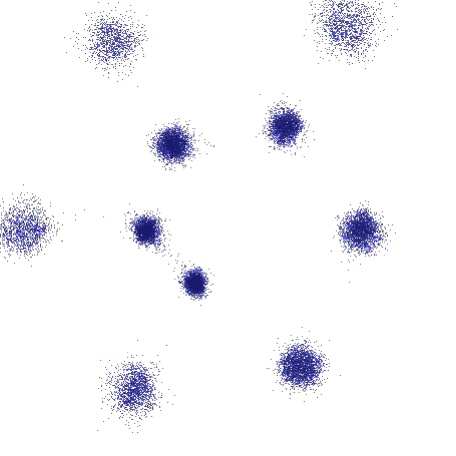};
\insertplotrow{PR=90\%}{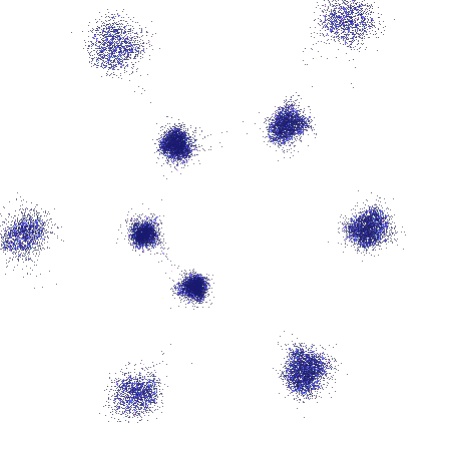};

\end{tikzpicture}
\caption{Pruning FFJORD (PR= Prune ratio).}
\label{fig:toy_flows}
\end{wrapfigure}

\subsection{Density Estimation on 2D Data}
\label{sec:experiments_2d}
In the first set of experiments, we train FFJORD on a multi-modal Gaussian distribution, a multi-model set of Gaussian distributions placed orderly on a spiral as well as a spiral distribution with sparse regions. Figure \ref{fig:toy_flows} (first row) illustrates that densely connected flows (prune ratio = 0\%) might get stuck in sharp local minima and as a result induce mode collapse \citep{NIPS2017_44a2e080}. Once we perform unstructured pruning, we observe that the quality of the density estimation in all tasks considerably improves, c.f. Figure \ref{fig:toy_flows} (second and third rows). If we continue sparsifying the flows, depending on the task at hand, the flows get disrupted again.

Therefore, there is a certain threshold for pruning flows required to avoid generative modeling issues such as mode-collapse in continuous flows. We validate this observation by plotting the negative log-likelihood loss as a function of the prune ratio in all three tasks with both unstructured and structured pruning. As shown in Figure \ref{fig:toy_nll}, we confirm that sparsity in flows improves the performance of continuous normalizing flows.

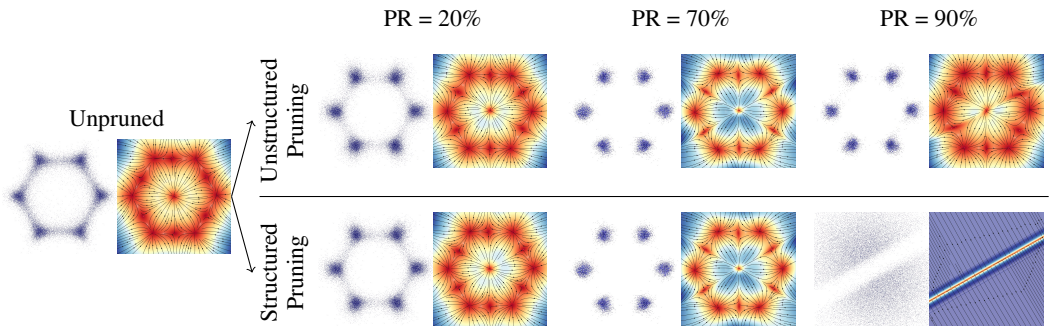
\begin{figure}[t]
\begin{tikzpicture}[scale=1.0]
\footnotesize
\newcommand{\tikzfigheight}{43pt}
\coordinate (top) at (0.0,2.1);
\coordinate (middle) at (0.0,0.0);
\coordinate (bottom) at (0.0,-2.1);
\coordinate (pr0) at (0.9,0);
\coordinate (prdiff) at (3.3,0);
\coordinate (textoffset) at (0,0.05);
\coordinate (picoffset) at (0,-0.2);
\coordinate (describeoffset) at (-2.4,-0.15);
\newcommand{\insertplotpair}[3]{
\node[below left, inner sep=0pt]  at ($({#1})+(pr0)+{#2}*(prdiff)+(picoffset)$) {\includegraphics[height=\tikzfigheight]{fig/ToyGaussians/ffjord_l2_h128_sigmoid_da_ToyGaussians_e100/distribution/#3.jpg}};
\node[below right, inner sep=0pt] at ($({#1})+(pr0)+{#2}*(prdiff)+(picoffset)$) {\includegraphics[height=\tikzfigheight]{fig/ToyGaussians/ffjord_l2_h128_sigmoid_da_ToyGaussians_e100/field/#3.jpg}};
}
\node[left, inner sep=0pt] (ref1)  at ($(middle)$) {\includegraphics[height=\tikzfigheight]{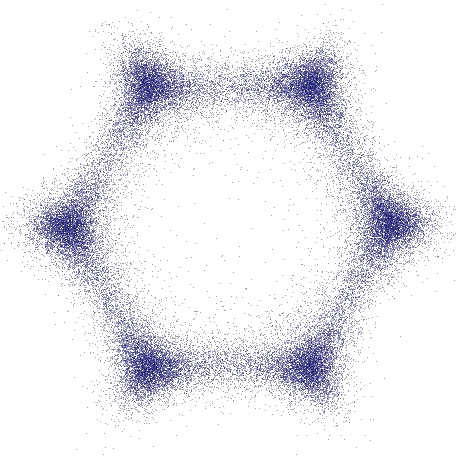}};
\node[right, inner sep=0pt] (ref2) at (ref1.east) {\includegraphics[height=\tikzfigheight]{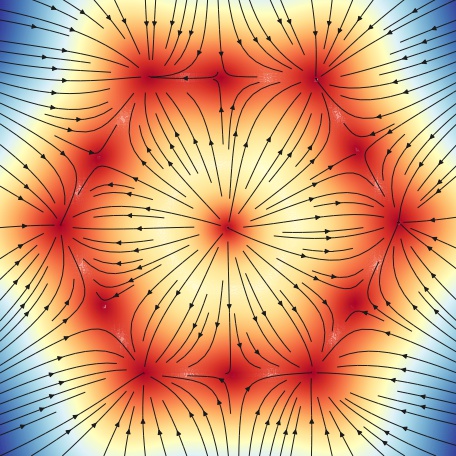}};
\insertplotpair{top}{1}{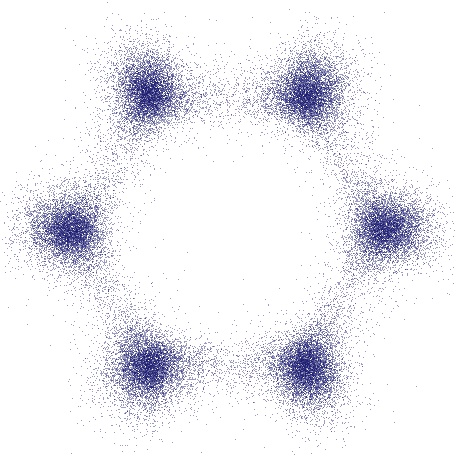}
\insertplotpair{top}{2}{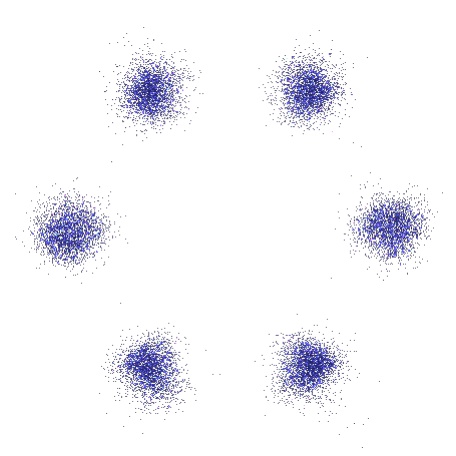}
\insertplotpair{top}{3}{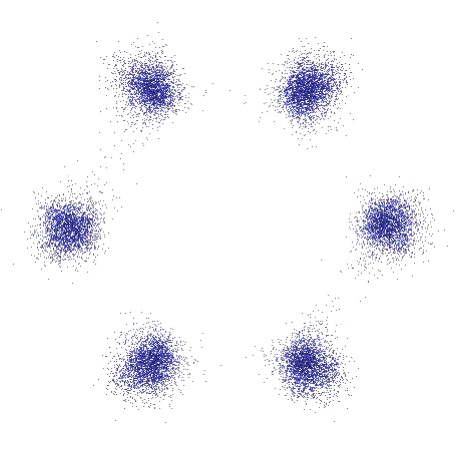}
\insertplotpair{middle}{1}{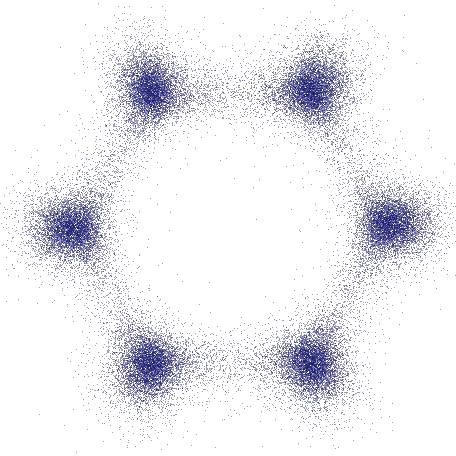}
\insertplotpair{middle}{2}{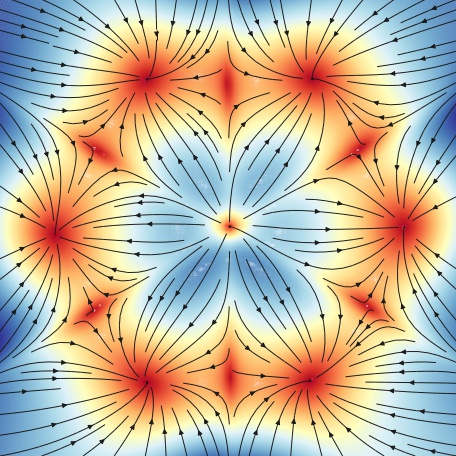}
\insertplotpair{middle}{3}{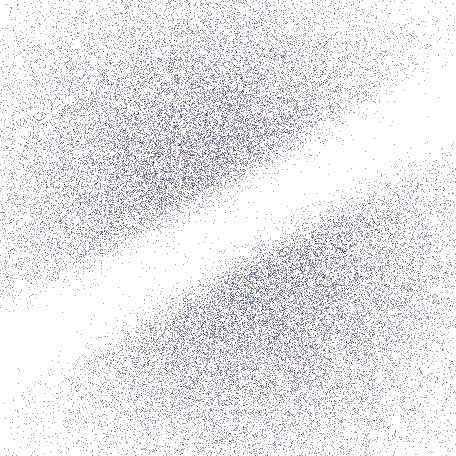}
\node[above] at (ref1.north east) {Unpruned};
\node[above] at ($(top)+(pr0)+1*(prdiff)+(textoffset)$) {PR = 20\%};
\node[above] at ($(top)+(pr0)+2*(prdiff)+(textoffset)$) {PR = 70\%};
\node[above] at ($(top)+(pr0)+3*(prdiff)+(textoffset)$) {PR = 90\%};
\node [rotate=90, below left,align=center] (unstruc) at ($(top)+(pr0)+1*(prdiff)+(describeoffset)$) {Unstructured\\Pruning};
\node [rotate=90, below left,align=center] (struc) at ($(middle)+(pr0)+1*(prdiff)+(describeoffset)-(0,0.09)$) {Structured\\Pruning};
\draw  (54pt,0.0) -- (\textwidth-45pt,0);
\draw [->] (ref2.east) -- (unstruc.north);
\draw [->] (ref2.east) -- (struc.north);
\end{tikzpicture}
\caption{Multi-modal Gaussian flow and pruning. We observe that Sparse Flows attract the vector-field directions uniformly towards the mean of each Gaussian distribution, while an unpruned flow does not exploit this feature and contains converging vectors in between Gaussians. See Supplements Section S2, for a detailed explanation of these observations.}
\label{fig:multi-modal-toy-flow}
\end{figure}

We further explore the inner dynamics of the flows between unpruned and pruned networks on the multi-modal case, with the aim of understanding how pruning enhances density estimation performance. Figure \ref{fig:multi-modal-toy-flow} represents the vector-field constructed by each flow to model 6-Gaussians independently. We observe that sparse flows with PR of 70\% attract the vector-field directions uniformly towards the mean of each Gaussian. In contrast, unpruned flows do not exploit this feature and contain converging vectors in between distributions. This is how the mode-collapse occurs. 

\subsection{Density Estimation on Real Data - Tabular}
We scale our experiments to a set of five real-world tabular datasets (prepared based on the instructions given by \citet{papamakarios2017masked} and \citet{grathwohl2019ffjord}) to verify our empirical observations about the effect of pruning on the generalizability of continuous normalizing flows. Table \ref{tab:tabular_results} summarizes the results. We observe that sparsifying FFJORD flows substantially improves their performance in all 5 tasks. In particular, we gain up to 42\% performance gain in the POWER, 35\% in GAS, 12\% in HEPMASS, 5\% in MINIBOONE and 19\% in BSDS300. 

More importantly, this is achieved with flows with 1 to 3 orders of magnitude less parameters compared to other advanced flows. On MINIBOONE dataset for instance, we found a sparse flow with only 4\% of its original network that outperforms its densely-connected FFJORD flow. On MINIBOONE, Autoregressive flows (NAF) and sum-of-squares models (SOS) which outperform all other models possess 8.03 and 6.87 million parameters. In contrast, we obtain a Sparse Flow with only 32K parameters that outperform all models except NAF and SOS. 

\begin{table}[t]
\setlength{\tabcolsep}{3.05pt} 
\renewcommand{\arraystretch}{1.2} 
\centering
\caption{Negative test log-likelihood (NLL) in nats of tabular datasets from~\citep{papamakarios2017masked} and corresponding architecture size in number of parameters (\#params). Sparse Flow (FFJORD with unstructured pruning) with lowest NLL and competing baseline with lowest NLL are bolded.}

\begin{adjustbox}{width=1\columnwidth}
\begin{tabular}{l|rr|rr|rr|rr|rr}
    \toprule
    \multirow{2}{*}{Model}
    & \multicolumn{2}{c|}{\textbf{\textsc{Power}}} 
    & \multicolumn{2}{c|}{\textbf{\textsc{Gas}}}
    & \multicolumn{2}{c|}{\textbf{\textsc{Hepmass}}}
    & \multicolumn{2}{c|}{\textbf{\textsc{Miniboone}}} 
    & \multicolumn{2}{c}{\textbf{\textsc{BSDS300}}} \\
    & nats & \#params
    & nats & \#params 
    & nats & \#params 
    & nats & \#params 
    & nats & \#params \\ 
    \hline
    MADE~{\tiny\citep{pmlr-v37-germain15}} 
    & 3.08 & 6K 
    & -3.56 & 6K 
    & 20.98 & 147K 
    & 15.59 & 164K 
    & -148.85 & 621K 
    \\
    Real NVP~{\tiny\citep{dinh2016density}}
    & -0.17 & 212K 
    & -8.33 & 216K 
    & 18.71 & 5.46M 
    & 13.84 & 5.68M 
    & -153.28 & 22.3M 
    \\
    MAF~{\tiny\citep{papamakarios2017masked}}
    & -0.24 & 59.0K 
    & -10.08 & 62.0K 
    & 17.70 & 1.47M 
    & 11.75 & 1.64M 
    & -155.69 &  6.21M 
    \\
    Glow~{\tiny\citep{kingma2018glow}}
    & -0.17 &  \na 
    & -8.15 & \na 
    & 18.92 & \na 
    & 11.35 & \na 
    & -155.07 & \na 
    \\
    CP-Flow~{\tiny\citep{huang2020convex}}
    & -0.52 & 5.46M  
    & -10.36 & 2.76M 
    & 16.93 & 2.92M  
    & 10.58 & 379K   
    & -154.99 & 2.15M 
    \\
    TAN~{\tiny\citep{pmlr-v80-oliva18a}}
    & \textbf{-0.60} &   \na
    & \textbf{-12.06} & \na
    & \textbf{13.78} & \na
    & 11.01 &  \na
    & \textbf{-159.80} &  \na
    \\
    NAF~{\tiny\citep{pmlr-v80-huang18d}}
    & \textbf{-0.62} &  \textbf{451K} 
    & \textbf{-11.96} & \textbf{443K} 
    & 15.09 & 10.7M 
    & \textbf{8.86} & \textbf{8.03M} 
    & -157.73 &  42.3M 
    \\
    SOS~{\tiny\citep{jaini2019sum}}
    & \textbf{-0.60}  & \textbf{212K} 
    & \textbf{-11.99} &  \textbf{256K} 
    & 15.15 &  4.43M 
    & \textbf{8.90}  & \textbf{6.87M} 
    & -157.48 &  9.09M 
    \\
    \hline
    FFJORD~{\tiny\citep{grathwohl2019ffjord}}
    & -0.35 & 43.3K 
    & -8.58 & 279K 
    & 17.53 & 547K 
    & 10.50 & 821K 
    & -128.33 & 6.70M 
    \\
    \hline
    \multirow{4}{*}{Sparse Flow}
    & -0.45 & 30K 
    & -10.79 & 194K 
    & 16.53 & 340K 
    & 10.84 & 397K  
    & -145.62 & 4.69M 
    \\
    & -0.50 & 23K 
    & -11.19 & 147K 
    & 15.82 & 160K 
    & 10.81 & 186K  
    & -148.72 & 3.55M 
    \\
    & \textbf{-0.53} & \textbf{13K} 
    & \textbf{-11.59} & \textbf{85K} 
    & \textbf{15.60} & \textbf{75K} 
    & \textbf{9.95} & \textbf{32K} 
    & -150.45 & 2.03M 
    \\
    & -0.52 & 10K 
    & -11.47 & 64K 
    & 15.99 & 46K 
    & 10.54 & 18K 
    & \textbf{-151.34} & \textbf{1.16M} 
    \\
    \bottomrule

\end{tabular}
\end{adjustbox}
\label{tab:tabular_results}
\end{table}

\begin{figure}[t!]
\centering
\begin{minipage}[t]{0.28\textwidth}
    \includegraphics[width=\textwidth]{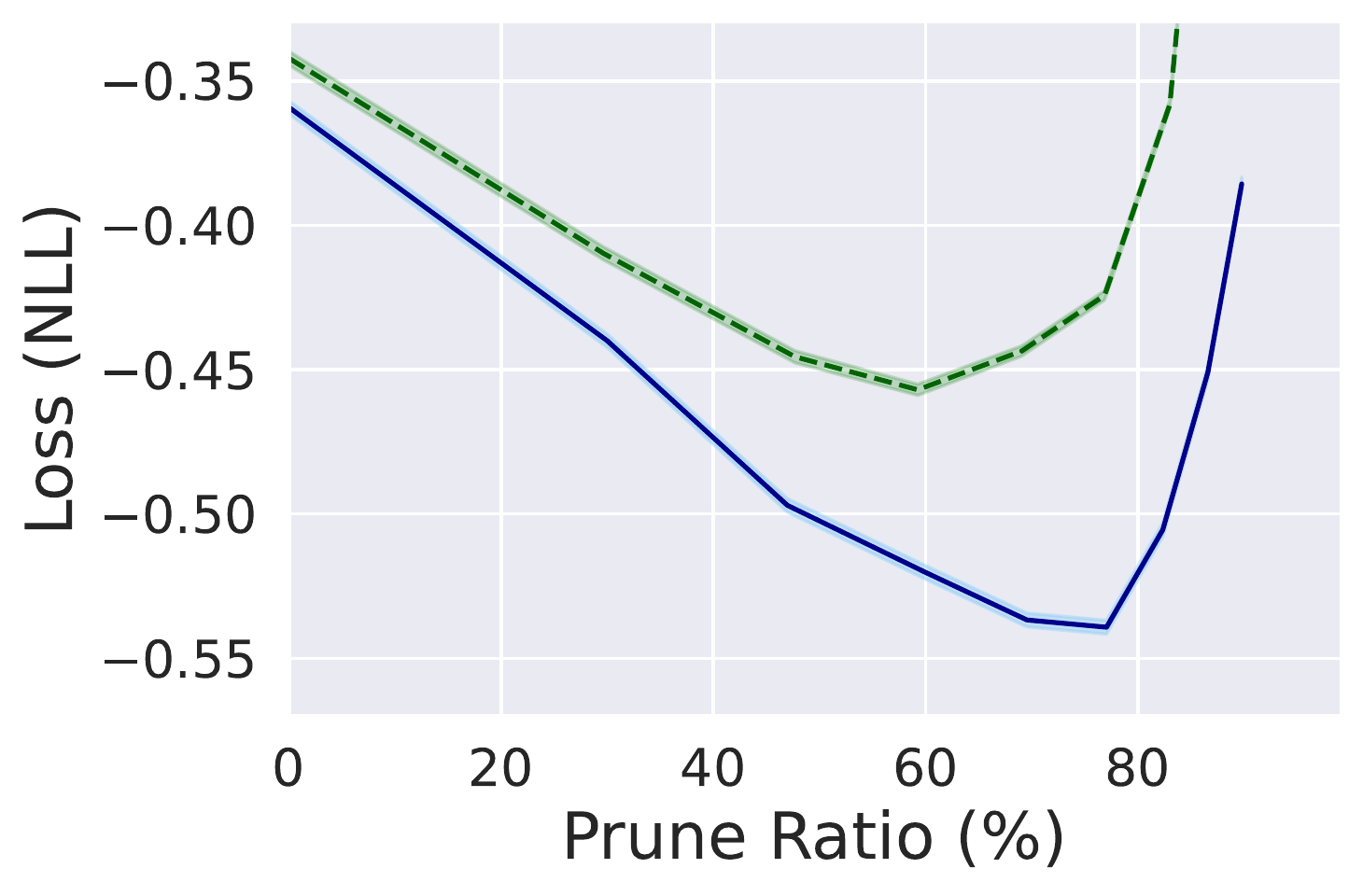}
    \subcaption{Power}
\end{minipage}%
\begin{minipage}[t]{0.28\textwidth}
    \includegraphics[width=\textwidth]{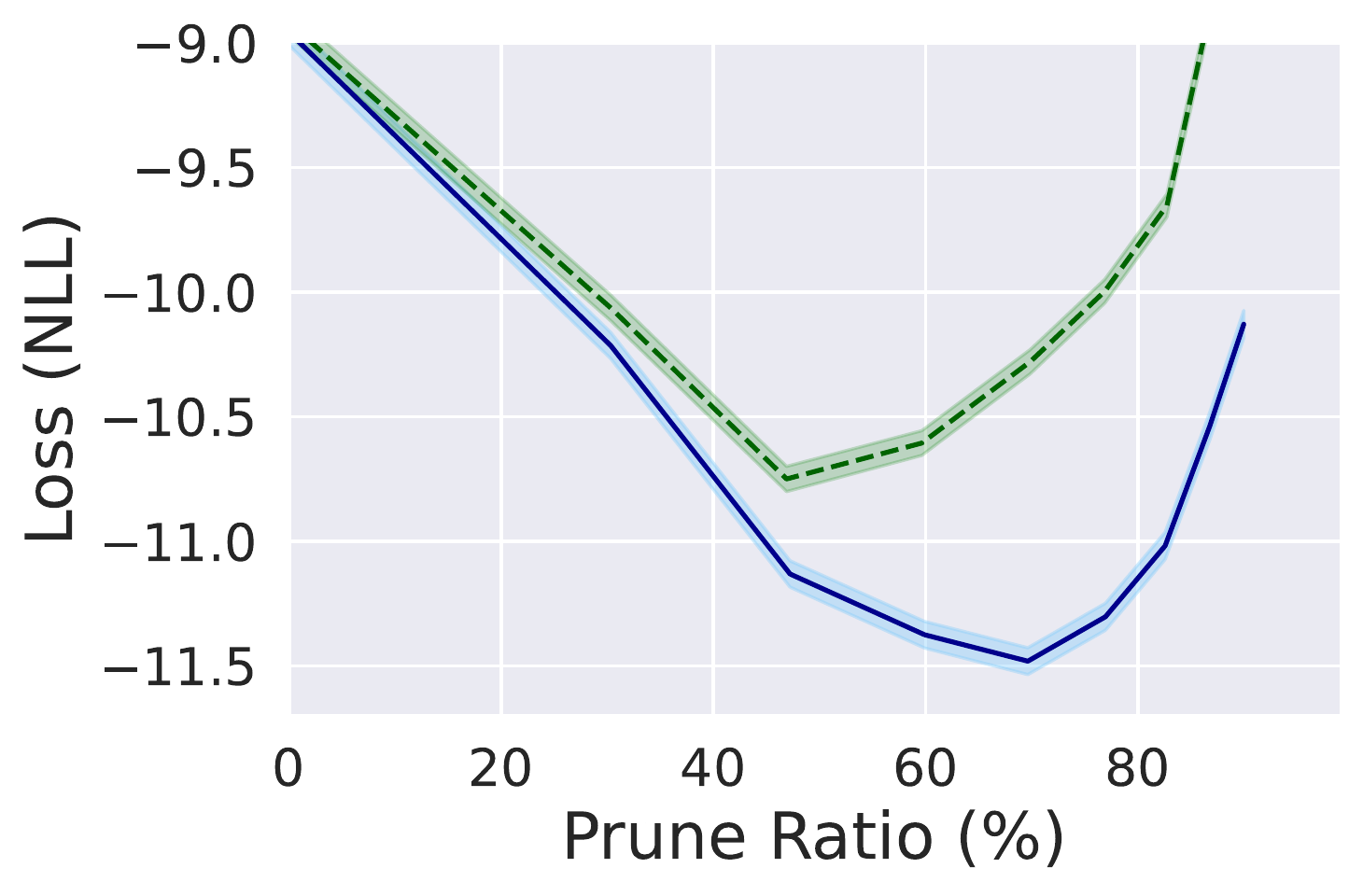} 
    \subcaption{Gas}
\end{minipage}%
\begin{minipage}[t]{0.28\textwidth}
    \includegraphics[width=\textwidth]{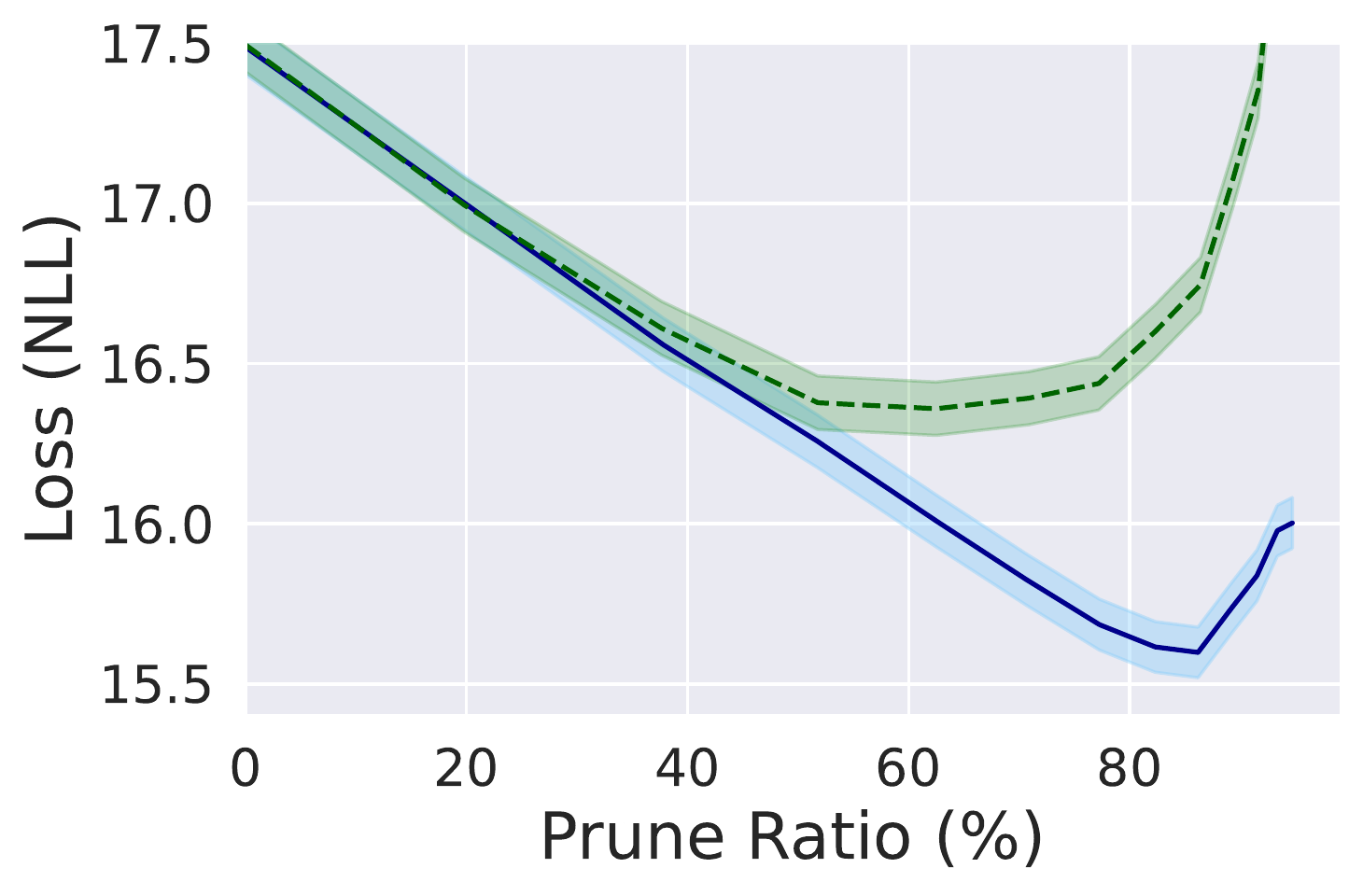}
    \subcaption{Hepmass}
\end{minipage}
\begin{minipage}[t]{0.28\textwidth}
    \includegraphics[width=\textwidth]{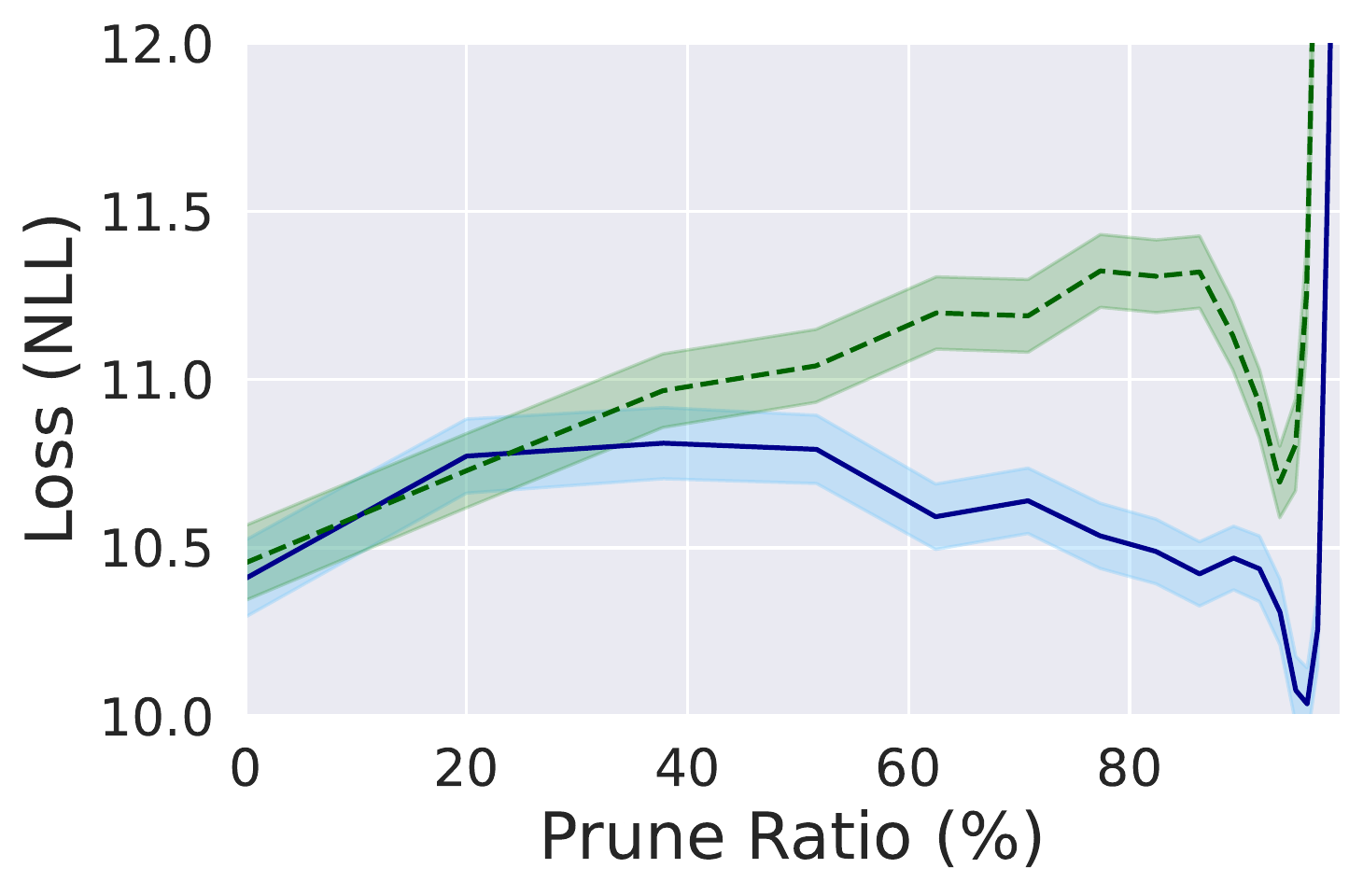}
    \subcaption{Miniboone}
\end{minipage}%
\begin{minipage}[t]{0.28\textwidth}
    \includegraphics[width=\textwidth]{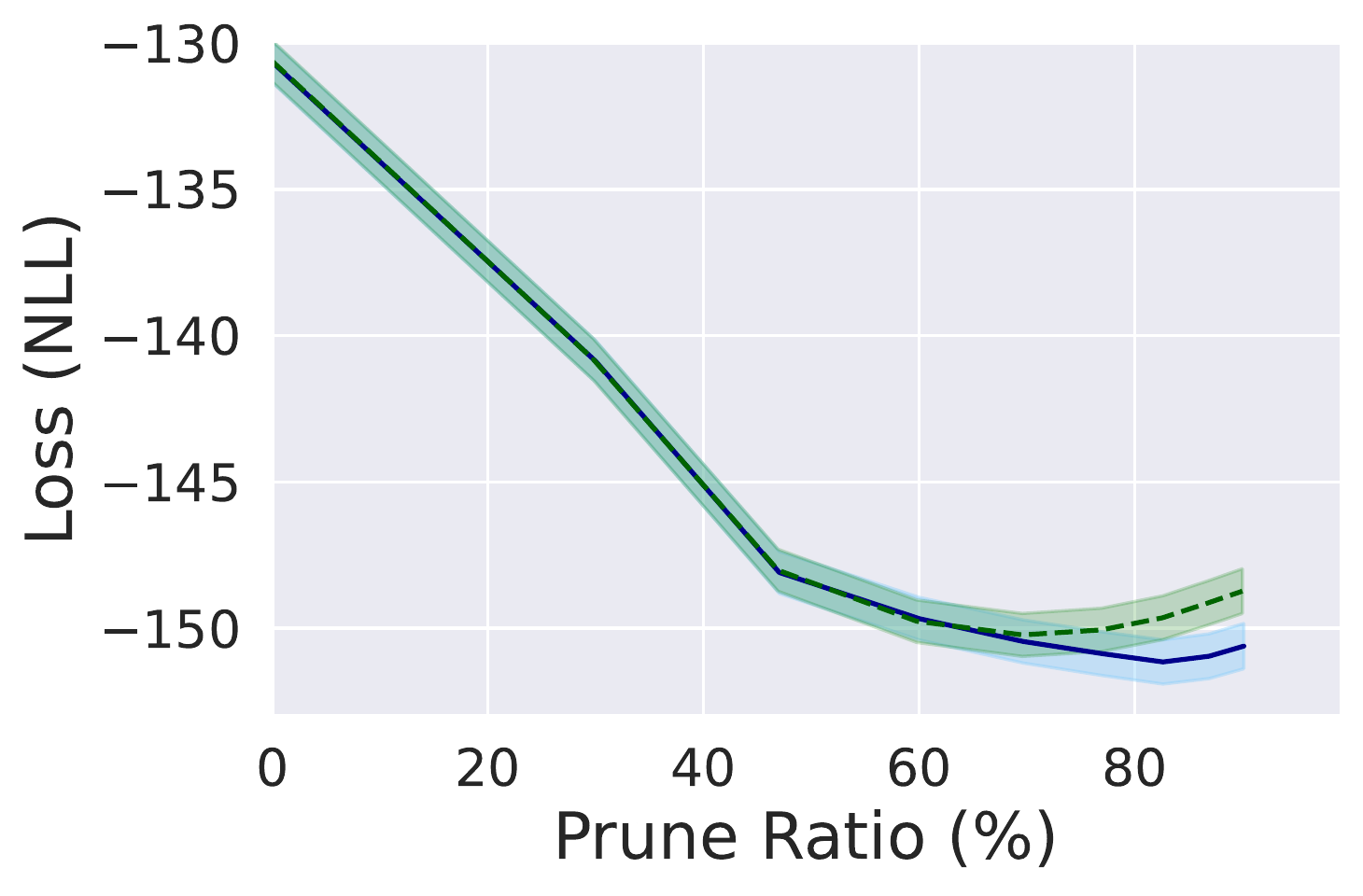}
    \subcaption{Bsds300}
\end{minipage}%
\begin{minipage}[t]{0.28\textwidth}
    \centering
    \vspace{-14ex}
    \hspace{2ex}
    \includegraphics[width=0.6\textwidth]{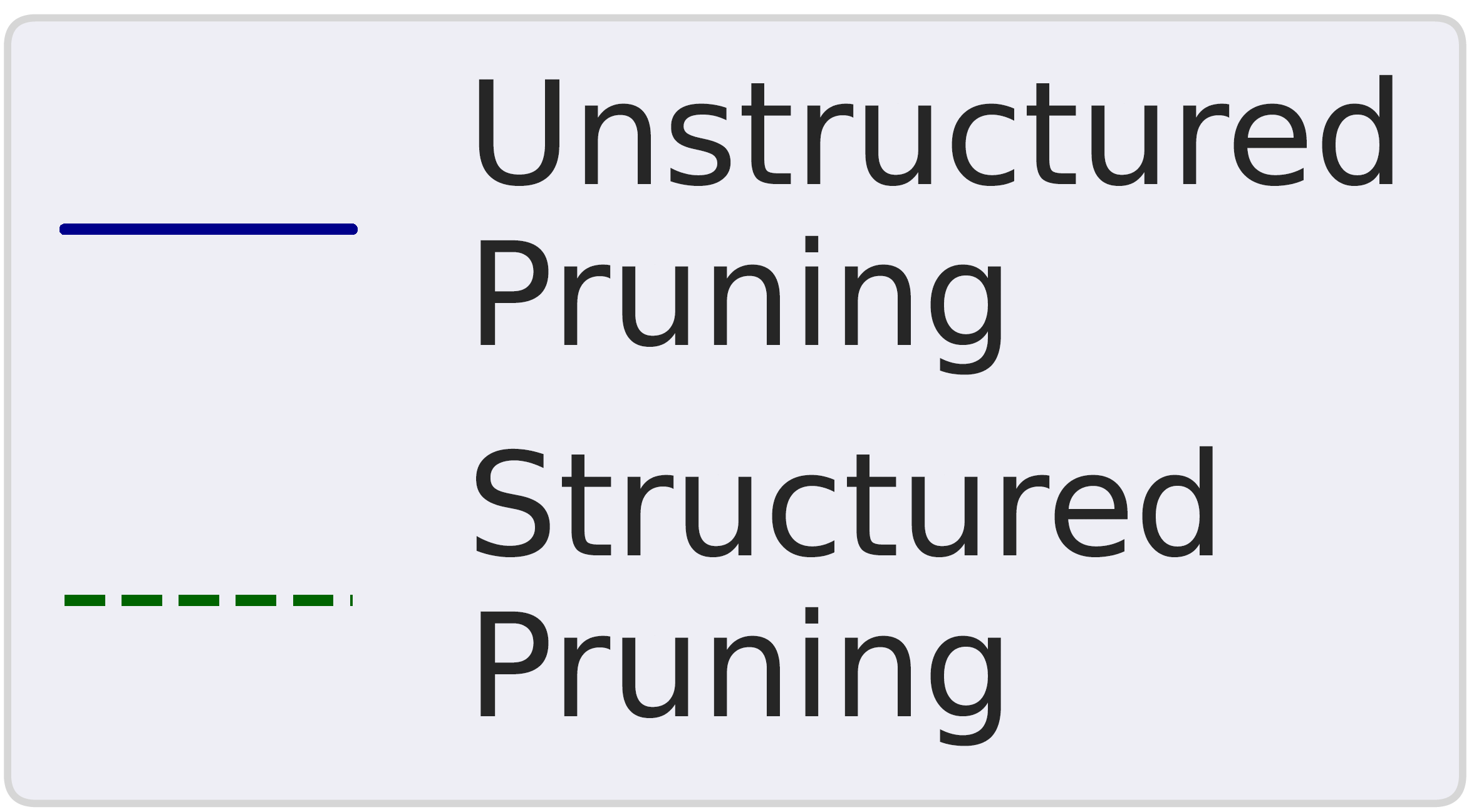}
\end{minipage}%
\caption{Negative log-likelihood versus prune ratio on tabular datasets.}
\label{fig:tabular-trends}
\end{figure}

Let us now look at the loss versus prune-ratio trends in all experiments to conclude our empirical observations on real-world tabular datasets. As shown in Figure \ref{fig:tabular-trends}, we observe that pruning considerably improves the performance of flows at larger scale as well. 

\begin{table}[t]
\renewcommand{\arraystretch}{1.15} 
\centering
\small
\caption{Negative test log-likelihood (NLL) in bits/dim for image datasets and corresponding architecture size in number of parameters (\#params). Sparse Flow (FFJORD with unstructured pruning) with lowest NLL and competing baseline with lowest NLL are bolded.}
\begin{tabular}{l|rr|rr}
    \toprule
    \multirow{2}{*}{Model}
    & \multicolumn{2}{c|}{\textbf{\textsc{MNIST}}} 
    & \multicolumn{2}{c}{\textbf{\textsc{CIFAR-10}}}\\
    & bits/dim & \#params 
    & bits/dim & \#params \\ 
    \hline
    MADE~\citep{pmlr-v37-germain15} 
    & 1.41 & 1.20M 
    & 5.80 & 11.5M 
    \\
    Real NVP~\citep{dinh2016density} 
    & 1.05 & \na
    & 3.49 & \na
    \\
    MAF~\citep{papamakarios2017masked} 
    & 1.91 & 12.0M 
    & 4.31 & 115M 
    \\
    Glow~\citep{kingma2018glow} 
    & 1.06 & \na
    & 3.35 & 44.0M
    \\
    CP-Flow~\citep{huang2020convex} 
    & 1.02 & 2.90M 
    & 3.40 & 1.90M 
    \\
    TAN~\citep{pmlr-v80-oliva18a} 
    & 1.19 & \na
    & 3.98 & \na
    \\
    SOS~\citep{jaini2019sum}
    & 1.81 & 17.2M 
    &  4.18 & 67.1M 
    \\
    RQ-NSF~\citep{durkan2019neural} 
    & & \na
    & 3.38 & 11.8M
    \\
    Residual Flow~\citep{chen2019residual} 
    & \textbf{0.97} & \textbf{16.6M}
    & \textbf{3.28} & \textbf{25.2M}
    \\
    \hline
    FFJORD~\citep{grathwohl2019ffjord})
    & 1.01 & 801K 
    & 3.44 & 1.36M 
    \\
    \hline
    Sparse Flows (PR=20\%) 
    & 0.97 & 641K 
    & 3.38 & 1.09M 
    \\
    Sparse Flows (PR=38\%) 
    & 0.96 & 499K 
    & 3.37 & 845K 
    \\
    Sparse Flows (PR=52\%) 
    & 0.95 & 387K 
    & \textbf{3.36} & \textbf{657K} 
    \\
    Sparse Flows (PR=63\%) 
    & \textbf{0.95} & \textbf{302K} 
    & 3.37 & 510K 
    \\
    Sparse Flows (PR=71\%) 
    & 0.96 & 234K 
    & 3.38 & 395K 
    \\
    Sparse Flows (PR=77\%) 
    & 0.97 & 182K 
    & 3.39 & 308K 
    \\
    Sparse Flows (PR=82\%) 
    & 0.98 & 141K 
    & 3.40 & 239K 
    \\
    Sparse Flows (PR=86\%) 
    & 0.97  & 109K 
    & 3.42 & 186K 
    \\
    \bottomrule
\end{tabular}
\label{tab:image_results}
\end{table}

\subsection{Density Estimation on Real Data - Vision}
Next, we extend our experiments to density estimation for image datasets, MNIST and CIFAR10. We observe a similar case on generative modeling with both datasets, where pruned flows outperform densely-connected FFJORD flows. On MNIST, a sparse FFJORD flow with 63\% of its weights pruned outperforms all other benchmarks. Compared to the second best flow (Residual flow), our sparse flow contains 70x less parameters (234K vs 16.6M). On CIFAR10, we achieve the second best performance with over 38x less parameters compared to Residual flows which performs best. 

\begin{wrapfigure}[10]{r}{0.52\textwidth}
\vspace{-5mm}
\centering
\begin{minipage}[t]{0.25\textwidth}
    \includegraphics[width=\textwidth]{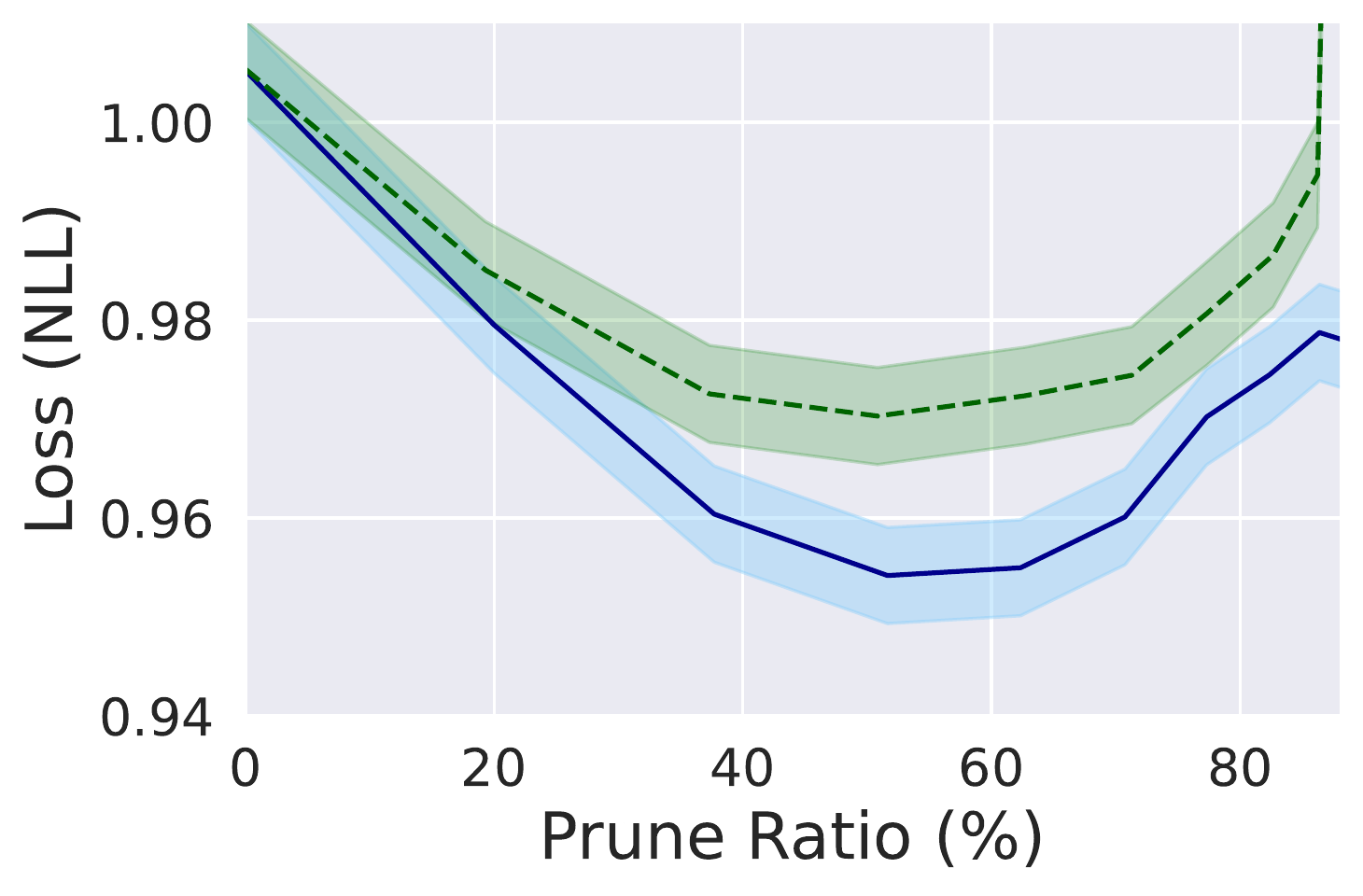}
    \subcaption{MNIST}
\end{minipage}%
\hspace{1ex}
\begin{minipage}[t]{0.25\textwidth}
    \includegraphics[width=\textwidth]{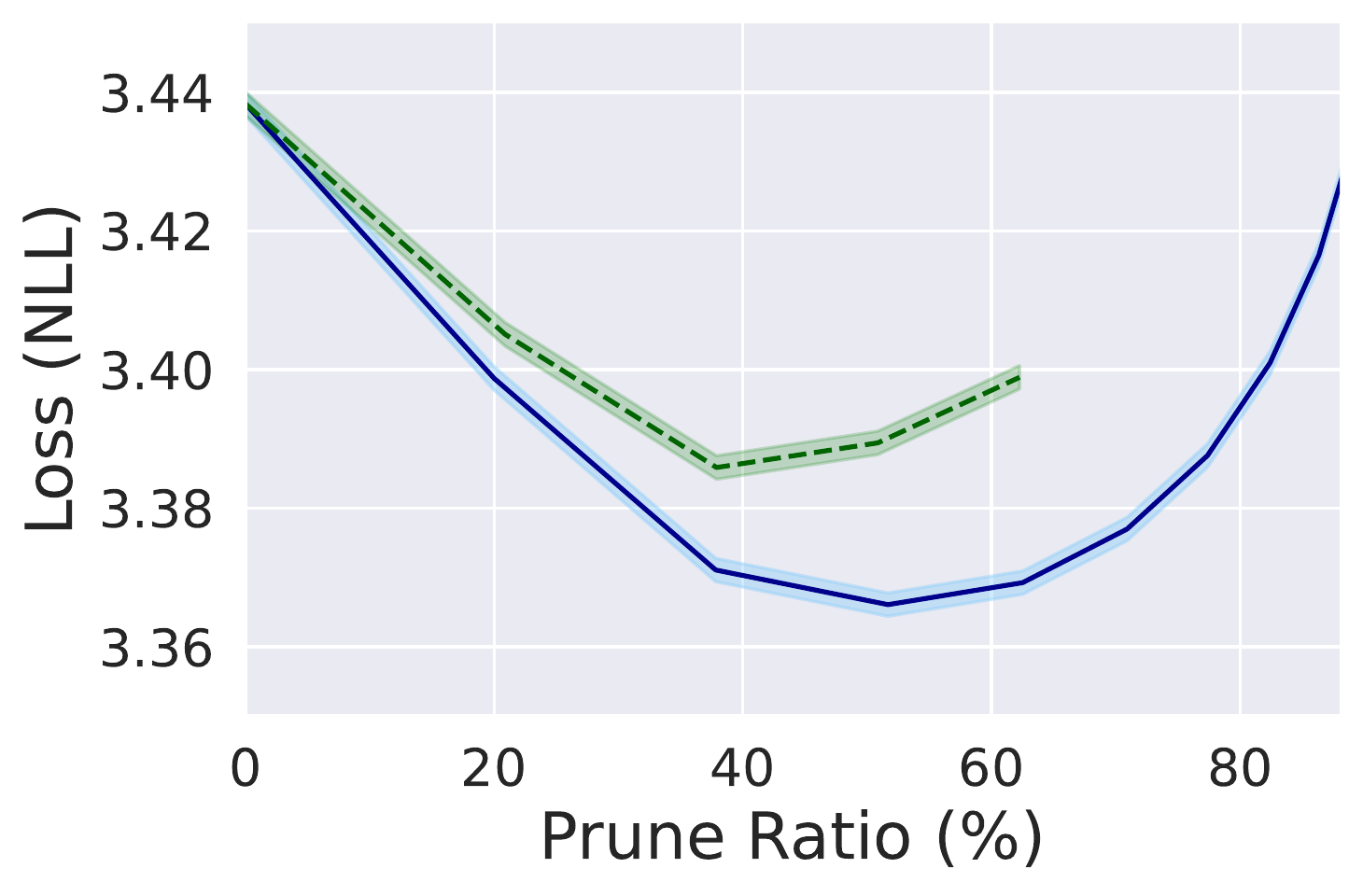}
    \subcaption{CIFAR10}
\end{minipage}%
\caption{Loss vs. prune ratio for MNIST and CIFAR10 with unstructured and structured pruning.}
\label{fig:curve-images}
\end{wrapfigure}

Furthermore, on CIFAR10, Glow with 44 million parameters performs on par with our sparse FFJORD network with 657k parameters. It is worth mentioning that FFJORD obtains this results by using a simple Gaussian prior, while Glow takes advantage of learned base distribution \citep{grathwohl2019ffjord}. Figure \ref{fig:curve-images} illustrates the improvement of the loss (negative log-likelihood) in density estimation on image datasets as a result of pruning neural ODEs. Around 60\% sparsity of a continuous normalizing flow leads to better generative modeling compared to densely structured flows.

\subsection{Pruning Helps Avoid Mode-collapse}
To quantitatively explore the effectiveness of pruning on avoiding mode-collapse (as shown qualitatively in Figure \ref{fig:toy_flows}, we developed a measure adopted from \citep{srivastava2017veegan}: The percentage of Good Quality samples--- We draw samples from a trained normalizing flow and consider a sample of “good quality” if it is within $n$ ($n$ is typically chosen to be 2, 3, or 5) standard deviations from its nearest mode. We then can report the percentage of the good quality samples as a measure of how well the generative model captures modes. 

Table \ref{tab:mode_collapse} summarizes the results on some synthetic datasets and the large-scale MNIST dataset. We can see that the number of “good quality samples”, which stand for a quantitative metric for mode-collapse, improves consistently with a certain percentage of pruning in all experiments. This observation is very much aligned with our results on the generalization loss, which we will discuss in the following.

\begin{table}[t]
    \centering
    \caption{Percentage of good quality samples as a measure of mode-collapse. Pruning method: structured pruning.}
    \begin{tabular}{c|cc|cc|ccc}
        \toprule
        & \multicolumn{2}{c|}{\textbf{\textsc{Gaussian}}} & \multicolumn{2}{c|}{\textbf{\textsc{Gaussian-spiral}}} & \multicolumn{3}{c}{\textbf{\textsc{MNIST}}}\\
        \midrule
        & \multicolumn{7}{c}{Percentage of good quality samples as a measure of mode-collapse} \\
        \midrule
Prune ratio (\%) &	std = 2 & std = 3 & std = 2 & std = 3 & Prune ratio (\%) &	std = 3 & std = 5\\
\midrule
0.0	 &  72.89\%	& 89.34\% &	66.65\%	& 84.01\% & 0.0 &	6.94\% &	58.24\% \\
20.0 &	79.46\%	& 93.45\% &	74.29\% & 90.64\% & 20.0 &	8.02\% &	65.98\% \\
\textbf{25.6} &	\textbf{81.40}\%	& \textbf{94.90}\% & \textbf{78.22}\%	& \textbf{93.02}\% & \textbf{37.8} &	\textbf{9.35}\% &	\textbf{68.11}\% \\
\textbf{52.0} &	\textbf{84.30}\%	& \textbf{96.85}\% & \textbf{80.40}\% & \textbf{94.29}\% & \textbf{51.7} &	\textbf{8.70}\% &	\textbf{67.24}\% \\
\textbf{71.2} &	\textbf{81.83}\%	& \textbf{94.86}\% & \textbf{78.06}\% & \textbf{93.41}\% & \textbf{77.3} &	\textbf{8.59}\% &	\textbf{66.24}\% \\
75.1 &	68.20\%	& 86.10\% & 78.99\% & 93.67\% & 89.3 &	6.07\% &	57.28\% \\
98.0 &	36.01\%	& 76.54\% & 9.27\%  & 20.32\% & 91.7 &	0.02\% &	32.75\% \\
         \bottomrule
    \end{tabular}
    \label{tab:mode_collapse}
\end{table}

\subsection{Pruning Flattens the Loss Surface for Neural ODEs}
What could be the potential reason for the enhanced generalization of pruned CNFs besides their ability to resolve mode-collapse? To investigate this further, we conducted a Hessian-based analysis on the flows-under-test. Dissecting the properties of the Hessian by Eigenanalysis allows us to gain useful insights about the behavior of neural networks \citep{hochreiter1997long,sagun2017empirical,ghorbani2019investigation,erichson2021lipschitz,lechner2020learning}. We use PyHessian \citep{yao2020pyhessian} tool-set to analyze the Hessian $H$ w.r.t. the parameters of the CNF. This enables us to study the curvature of the loss function as the eigenvalues of the Hessian determines (locally) the loss gradients' changes. 

\begin{table}[b]
	\caption{Eigenanalysis of the Hessian $H$ in terms of the largest eigenvalue ($\lambda_{max}$), trace ($\text{tr}$), and condition number ($\kappa$) of pruned and unpruned continuous normalizing flows on the mixture of Gaussian task. Numbers are normalized with respect to the unpruned flow. See more results on other datasets in the supplements Section S3.}
	\centering
	\begin{tabular}{lcccc}
		\toprule
		Model & NLL & $\lambda_{max}(H)$ & $\text{tr}(H)$ & $\kappa(H)$ \\
		\midrule
Unpruned FFJORD        & 1.309 & 1.000 & 1.000 & 1.000 \\
Sparse Flows (PR=20\%) & 1.163 & 0.976 & 0.858 & 0.825 \\
Sparse Flows (PR=60\%) & 1.125 & 0.356 & 0.583 & 0.717 \\
Sparse Flows (PR=70\%) & \textbf{1.118} & \textbf{0.295} & \textbf{0.340} & \textbf{0.709} \\
Sparse Flows(PR=90\%) & 1.148 & 0.416 & 0.366 & 0.547 \\
\bottomrule
	\end{tabular}
	\label{tab:hessian}
\end{table}

Larger Hessian eigenvalues therefore, stand for sharper curvatures and their sign identifies upward or downward curvatures. In Table \ref{tab:hessian}, we report the maximum eigenvalue of the Hessian $\lambda_{max}$, Hessian's Trace, and the condition number $\kappa = \frac{\lambda_{max}}{\lambda_{min}}$.\footnote{This experiment was inspired by the Hessian-based robustness analysis performed in \cite{erichson2021lipschitz}.} Smaller $\lambda_{max}$ and tr($H$) indicates that our normalizing flow found a flatter minimum. As shown in Figure \ref{fig:flas-vs-sharp-minimum}, a flat minimum leads to a better generalization error as opposed to a sharp minimum. We find that up to a certain prune ratio, the maximum of the Hessian decreases and increases thereafter. Therefore, we claim that pruned continuous flows finds flatter local minimum, therefore, it generalize better than their unpruned version. 

Moreover, the Hessian condition number $\kappa$ could be an indicator of the robustness and efficiency of a deep model \citep{NIPS2007_0d3180d6}. Smaller $\kappa$ corresponds to obtaining a more robust learned agent. We observe that $\kappa$ also follows the same trend and up to a certain prune ratio it shrinks and then increases again confirming our hypothesis. 

\subsection{On the Robustness of Decision Boundaries of Pruned ODE-based Flows}

While pruning improves performance, it is imperative to investigate their robustness properties in constructing decision boundaries \citep{lechner2021adversarial}. For feedforward deep models it was recently shown that although pruned networks perform as well as their unpruned version, their robustness significantly decreases due to smaller network size \citep{liebenwein2021lost}. Is this also the case for neural ODEs? 

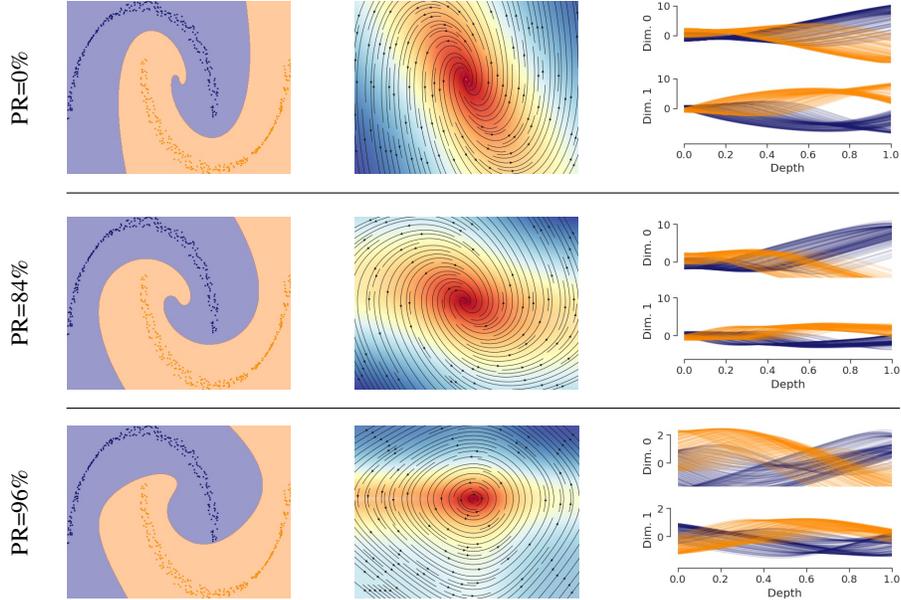
\begin{figure}[t!]
\centering
\begin{tikzpicture}[scale=1.0]

\small

\newcommand{\tikzfigheight}{65pt}

\coordinate (sepstart) at (0.0,2.1);

\coordinate (textoffset) at (0.4,0);
\coordinate (picoffset) at (0.85,0);

\coordinate (lineoffset) at (0.0, -0.25);

\newcommand{\insertplotrow}[2]{
\node [rotate=90, below left,align=center] (prtext) at ($(sepstart)$) {#1};

\node[right, inner sep=0pt] (plot1) at ($(prtext.south) + (textoffset)$) {\includegraphics[height=\tikzfigheight]{fig/ToyMoons/node_l2_h128_tanh_da_ToyMoons_e50/distribution_original/#2.jpg}};

\node[right, inner sep=0pt] (plot2) at ($(plot1.east) + (picoffset)$) {\includegraphics[height=\tikzfigheight]{fig/ToyMoons/node_l2_h128_tanh_da_ToyMoons_e50/field/#2.jpg}};

\node[right, inner sep=0pt] (plot3) at ($(plot2.east) + (picoffset)$) {\includegraphics[height=\tikzfigheight]{fig/ToyMoons/node_l2_h128_tanh_da_ToyMoons_e50/trajectory/#2.jpg}};

\coordinate (sepstart) at ($(sepstart) - (0,\tikzfigheight) + 2*(lineoffset)$);
}

\newcommand{\drawsepline}{
\draw ($(plot1.south west) + (lineoffset)$) -- ($(plot3.south east) + (lineoffset)$);
}

\insertplotrow{PR=0\%}{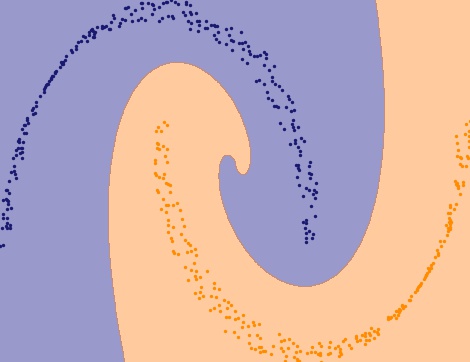};
\drawsepline;
\insertplotrow{PR=84\%}{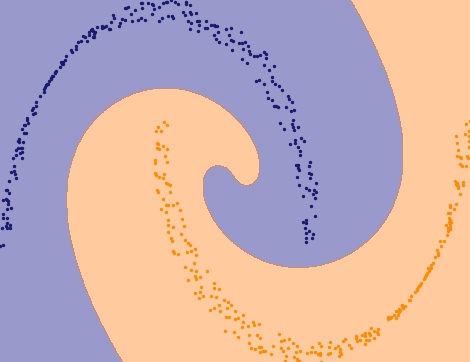};
\drawsepline;
\insertplotrow{PR=96\%}{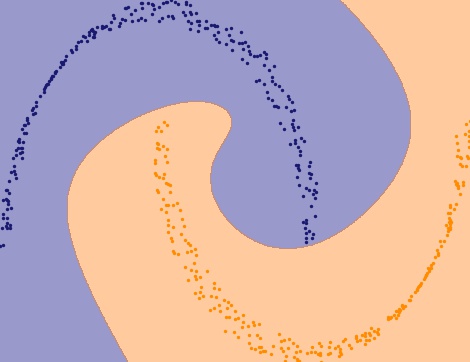};

\end{tikzpicture}
\caption{Robustness of decision boundaries for pruned networks. Column 1 is the decision boundary. Column 2 = state-space, and column 3 = the flow of data points: it shows the evolution of single samples (each narrow line represents an individual sample’s dimension 1 path) from input to their output through the depth of the network. How does the attenuation of the vector field intensity indicate that the classification boundary is more sensitive to perturbations? Let’s look at the first column and row 3 entry: We see that the classification boundary between the blue half-moon and the orange half-moon is very close to one tail of the blue half-moon. This means that if we perturb the blue half-moon data, some of them might get classified incorrectly based on the extremely pruned network’s decision boundary. As the intensity of the vector field shrinks to the center of the vector field (it is attenuated in the areas distant from the center), the decision boundary is set poorly as opposed to networks with more parameters.}
\label{fig:classification}
\end{figure}

We design an experiment to investigate this. We take a simple 2-Dimensional moon dataset and perform classification using unpruned and pruned neural ODEs.\footnote{This experiment is performed using the TorchDyn library \citep{poli2020torchdyn}.} This experiment (shown in Figure \ref{fig:classification}) demonstrates another valuable property of neural ODEs: We observe that neural ODE instances pruned up to 84\%, manage to establish a safe decision boundary for the two half moon classes. This insight about neural ODEs which was obtained by our pruning framework is another testimony of the merits of using neural ODE in decision-critical applications. 
Additionally, we observe that the decision-boundary gets very close to one of the classes for a network pruned up to 96\% which reduces robustness. Though even this network instance provide the same classification accuracy compared to the densely connected version. 

The state-space plot illustrates that a neural ODE's learned vector-field becomes edgier as we keep pruning the network. Nonetheless, we observe that the distribution of the vector-field's intensity (the color map in the second column) get attenuated as we go further away from its center, in highly pruned networks. This indicates that classification near the decision boundary is more sensitive to perturbations in extremely pruned networks. 

Moreover, Column 3 of Figure \ref{fig:classification} shows that when the networks are not pruned (column 3, row 1) or are pruned up to a certain threshold (column 3, row 2), the separation between the flow of samples of each class (orange or blue) is more consistent compared to the very pruned network (column 3, row 3). In Column 3, row 3 we see that the flow of individual samples from each class is more dispersed. As a result, the model is more sensitive to perturbations on the input space. This is very much aligned with our observation from column 1 and column 2 of Figure \ref{fig:classification}.

\section{Discussions, Scope and Conclusions}
\label{sec:conclusions}

We showed the effectiveness of pruning for continuous neural networks. Pruning improved generalization performance of continuous normalizing flows in density estimation tasks at scale. Additionally, pruning allows us to obtain performant minimal network instances with at least one order of magnitude less parameter count. By providing key insights about how pruning improves generative modeling and inference, we enabled the design of better neural ODE instances. 

\textbf{Ensuring sparse Hessian computation for sparse continuous-depth models.} As we prune neural ODE instances, their weight matrices will contain zero entries. However the Hessian with respect to those zero entries is not necessarily zero. Therefore, when we compute the eigenvalues of the Hessian, we must ensure to make the decompositon vector sets the Hessian of the pruned weights to zero before performing our eigenanalysis.

\textbf{Why experimenting with FFJORD and not CNFs?} FFJORD is an elegant trick to efficiently compute the log determinant of the Jacobians in the change of variables formula for CNFs. Thus, FFJORDs are CNFs but faster. To confirm the effectiveness and scalability of our approach, we pruned some CNFs on toy datasets and concluded very similar outcomes (see these results in Appendix).

\textbf{What are the limitations of Sparse Flows?} Similar to any ODE-based learning system, the computational efficiency of sparse flows is highly determined by the choice of their ODE solvers, data and model parameters. As the complexity of any of these fronts increases, the number of function evaluations for a given task increases. Thus we might have a slow training process. This computational overhead can be relaxed in principle by the use of efficient ODE solvers \citep{poli2020hypersolvers} and flow regularization schemes \citep{massaroli2020dissecting}.

\textbf{What design notes did we learn from applying pruning to neural ODEs?} We performed an ablation study over different types of features of neural ODE models. These experiments can be found in the supplements Section S4. Our framework suggested that to obtain a generalizable sparse neural ODE representation, the choice of activation function is important. In particular activations that are Lipschitz continuous, monotonous, and bounded are better design choices for density estimation tasks. 

Moreover, we found that the generalizability of sparse neural ODEs is more influenced by their neural network's width than their depth (number of layers), c.f., Appendix Section S4. Furthermore, pruning neural ODEs allows us to obtain better hyperparameters for the optimization problem by setting a trade-off between the value of the learning rate and weight decay.

In summary, we hope to have shown compelling evidence for the effectiveness of having sparsity in ODE-based flows.

\section*{Acknowledgments}
This research was sponsored by the United States Air Force Research Laboratory and the United States Air Force Artificial Intelligence Accelerator and was accomplished under Cooperative Agreement Number FA8750-19-2-1000. The views and conclusions contained in this document are those of the authors and should not be interpreted as representing the official policies, either expressed or implied, of the United States Air Force or the U.S. Government. The U.S. Government is authorized to reproduce and distribute reprints for Government purposes notwithstanding any copyright notation herein.
This work was further supported by The Boeing Company and the Office of Naval Research (ONR) Grant N00014-18-1-2830.


\bibliography{misc/references}


\clearpage
\newcommand{\beginsupplement}{%
        \setcounter{table}{0}
        \renewcommand{\thetable}{S\arabic{table}}%
        \setcounter{algorithm}{0}
        \renewcommand{\thealgorithm}{S\arabic{algorithm}}%
        \setcounter{equation}{0}
        \renewcommand{\theequation}{S\arabic{equation}}%
        \setcounter{figure}{0}
        \renewcommand{\thefigure}{S\arabic{figure}}%
        \setcounter{section}{0}
        \renewcommand{\thesection}{S\arabic{section}}%
     }

\setauthors


\title{Supplementary Material}
\maketitleextranoauthors

\beginsupplement

\vspace{-2ex}
\section{Hyperparameters}
We provide the necessary hyperparameters to reproduce our experiments below. For each set of experiments (Toy, Tabular, Images) we summarize the architecture of the unpruned model and relevant hyperparameters pertaining to the training/pruning process. All experiments were repeated \textbf{three times} with separate random seeds and average results are reported.

\begin{table}[h]
\centering
\caption{\textbf{Toy Dataset} Hyperparameters.}
\label{tab-supp:toy_hyperparameters}
\begin{adjustbox}{width=1.0\textwidth}
\begin{tabular}{c|l|cccc}
\toprule
 \multicolumn{2}{c|}{Hyperparameters}
& \textbf{\textsc{Gaussians}} & \textbf{\textsc{GaussianSpiral}} & \textbf{\textsc{Spiral}} & \textbf{\textsc{Moon}} \\ \midrule
\multirow{4}{*}{Architecture} 
& Layers 
& 2 & 4 & 4 & 2\\
& Hidden Size 
& 128 & 64 & 64 & 128 \\
& Activation
& Sigmoid & Sigmoid & Sigmoid & Tanh \\ 
& Divergence 
& Hutchison & Hutchison & Hutchison & Hutchison \\ \midrule
\multirow{4}{*}{Solver} 
& Type 
& Dopri & Dopri & Dopri & Dopri \\
& Rel. tol. 
& 1.0e-5 & 1.0e-5 & 1.0e-5 & 1.0e-4 \\
& Abs. tol 
& 1.0e-5 & 1.0e-5 & 1.0e-5 & 1.0e-4 \\
& Backprop.
& Adjoint & Adjoint & Adjoint & Adjoint \\ \midrule
\multirow{7}{*}{(Re-)Training} 
& Optimizer     
& AdamW  & AdamW & AdamW & Adam \\
& Epochs        
& 100 & 100 & 100 & 50 \\
& Batch size
& 1024 & 1024 & 1024 & 128 \\
& LR
& 5.0e-3 & 5.0e-2 & 5.0e-2 & 1.0e-2\\
& $\beta_1$ 
& 0.9 & 0.9 & 0.9 & 0.9 \\
& $\beta_2$ 
& 0.999 & 0.999 & 0.999 & 0.999 \\
& Weight decay  
& 1.0e-5 & 1.0e-2 & 1.0e-6 & 1.0e-4\\ \midrule
\multirow{1}{*}{Pruning}
& $PR$
& 10\% & 10\% & 10\% & 10\% \\  \bottomrule
\end{tabular}
\end{adjustbox}
\end{table}

\begin{table}[h]
\centering
\caption{\textbf{Tabular Datasets} Hyperparameters.}
\label{tab-supp:tabular_hyperparameters}
\begin{adjustbox}{width=1.0\textwidth}
\begin{tabular}{c|l|ccccc}
\toprule
 \multicolumn{2}{c|}{Hyperparameters}
& \textbf{\textsc{Power}} & \textbf{\textsc{Gas}} & \textbf{\textsc{Hepmass}} & \textbf{\textsc{Miniboone}} & \textbf{\textsc{Bsds300}} \\ \midrule
\multicolumn{2}{c|}{Architecture} 
& \multicolumn{5}{l}{Please refer to Table 4, Appendix B.1 of \citet{grathwohl2019ffjord}.} \\ \midrule
\multicolumn{2}{c|}{Solver} 
& \multicolumn{5}{l}{Please refer to Appendix C of \citet{grathwohl2019ffjord}.} \\ \midrule
\multirow{8}{*}{(Re-)Training} 
& Optimizer 
& Adam & Adam & Adam & Adam & Adam \\
& Epochs        
& 100 & 30 & 400 & 400 & 100 \\
& Batch size
& 10000 & 1000 & 10000 & 1000 & 10000 \\
& LR
& 1.0e-3 & 1.0e-3 & 1.0e-3 & 1.0e-3 & 1.0e-3 \\
& LR step
& 0.1@\{90, 97\} &  0.1@\{25, 28\} &  0.1@\{250, 295\} &  0.1@\{300, 350\} &  0.1@\{96, 99\} \\
& $\beta_1$ 
& 0.9 & 0.9 & 0.9 & 0.9 & 0.9 \\
& $\beta_2$ 
& 0.999 & 0.999 & 0.999 & 0.999 & 0.999 \\
& Weight decay  
& 1.0e-6 & 1.0e-6 & 1.0e-6 & 1.0e-6 & 1.0e-6 \\ \midrule
\multirow{1}{*}{Pruning}
& $PR$
& 25\% & 25\% & 22\% & 22\% & 25\% \\  \bottomrule
\end{tabular}
\end{adjustbox}
\end{table}

\begin{table}[H]
\centering
\caption{\textbf{Image Datasets} Hyperparameters.}
\label{tab-supp:image_hyperparameters}
\begin{adjustbox}{width=1.0\textwidth}
\begin{tabular}{c|l|cc}
\toprule
 \multicolumn{2}{c|}{Hyperparameters}
& \textbf{\textsc{MNIST}} & \textbf{\textsc{CIFAR-10}} \\ \midrule
\multicolumn{2}{c|}{Architecture} 
& \multicolumn{2}{l}{Please refer to Appendix B.1 (multi-scale) of \citet{grathwohl2019ffjord}.} \\ \midrule
\multicolumn{2}{c|}{Solver} 
& \multicolumn{2}{l}{Please refer to Appendix C of \citet{grathwohl2019ffjord}.} \\ \midrule
\multirow{8}{*}{(Re-)Training} 
& Optimizer 
& Adam & Adam \\
& Epochs        
& 50 & 50 \\
& Batch size
& 200 & 200 \\
& LR
& 1.0e-3 & 1.0e-3 \\
& LR step
& 0.1@\{45\} & 0.1@\{45\} \\
& $\beta_1$ 
& 0.9 & 0.9  \\
& $\beta_2$ 
& 0.999 & 0.999 \\
& Weight decay
& 0.0 & 0.0 \\ \midrule
\multirow{1}{*}{Pruning}
& $PR$
&  22\% &  22\% \\  \bottomrule
\end{tabular}
\end{adjustbox}
\end{table}

\section{Figure 6 Clarifications}
In figure below, we show the distribution and vector field of learned FFJORD networks (unpruned (top) and 70\% pruned down). The area in the vector field declared with a black circle shows the vector field structure around an actual mode in the dataset. We see that the vector field (which is illustrated by black arrows) attracts samples towards the mean of this distribution in both pruned and unpruned networks.

However, there is a drastic difference between the vector field structure in-between modes (annotated by purple circles), between the unpruned and pruned network. In the unpruned network, the vector field attracts samples in-between modes. In contrast, in the pruned network, the vector field is repellent in-between modes. Correspondingly, this illustration shows how an unpruned network tends to have samples in-between modes, while the pruned network avoids this shortcoming.

\begin{figure}[h]
	\centering
	\includegraphics[width=0.7\textwidth]{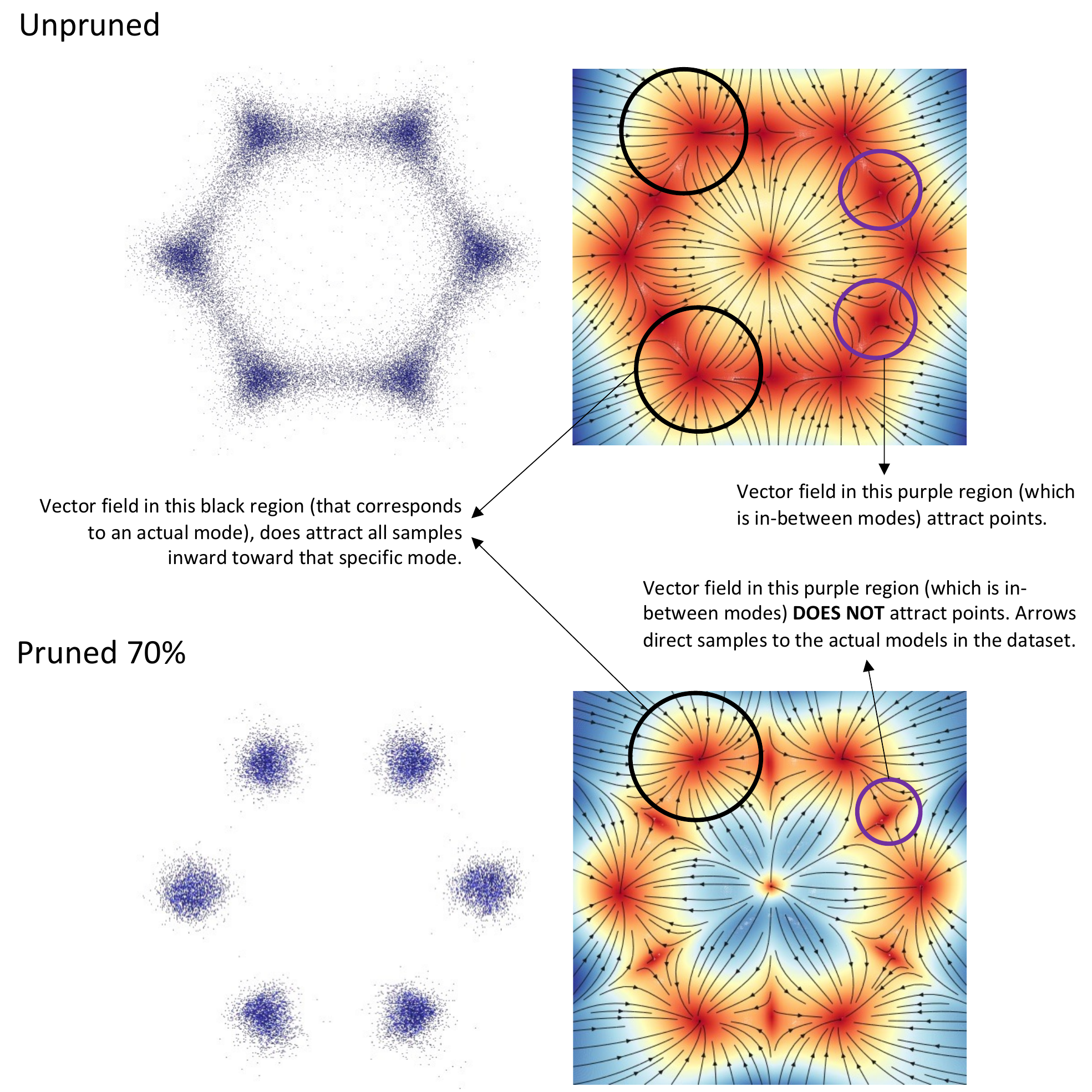}
	\caption{Clarification of how unpruned networks led to mode-collapse and pruned networks did not.}
	\label{fig:fig6_clarification}
\end{figure}

\section{Density Estimation on 2D Data with regular CNF}
FFJORD is an efficient way to reparameterize continuous normalizing flows (CNFs) during the backwards pass in order to avoid computing the full Hessian, which can be prohibitively expensive in large-scale experiments. In this section, we validate whether our observations hold independent of the method to compute the Hessian.

Specifically, we train regular CNFs with full Hessian computation on a multi-modal Gaussian distribution, a multi-model set of Gaussian distributions placed orderly on a spiral as well as a spiral distribution with sparse regions following the setup from Section~\ref{sec:experiments_2d}. 

\begin{figure}[H]
\centering
\begin{minipage}[t]{0.52\textwidth}
    \includegraphics[width=\textwidth]{fig/labels.pdf}
\end{minipage}
\begin{minipage}[t]{0.33\textwidth}
    \includegraphics[width=\textwidth]{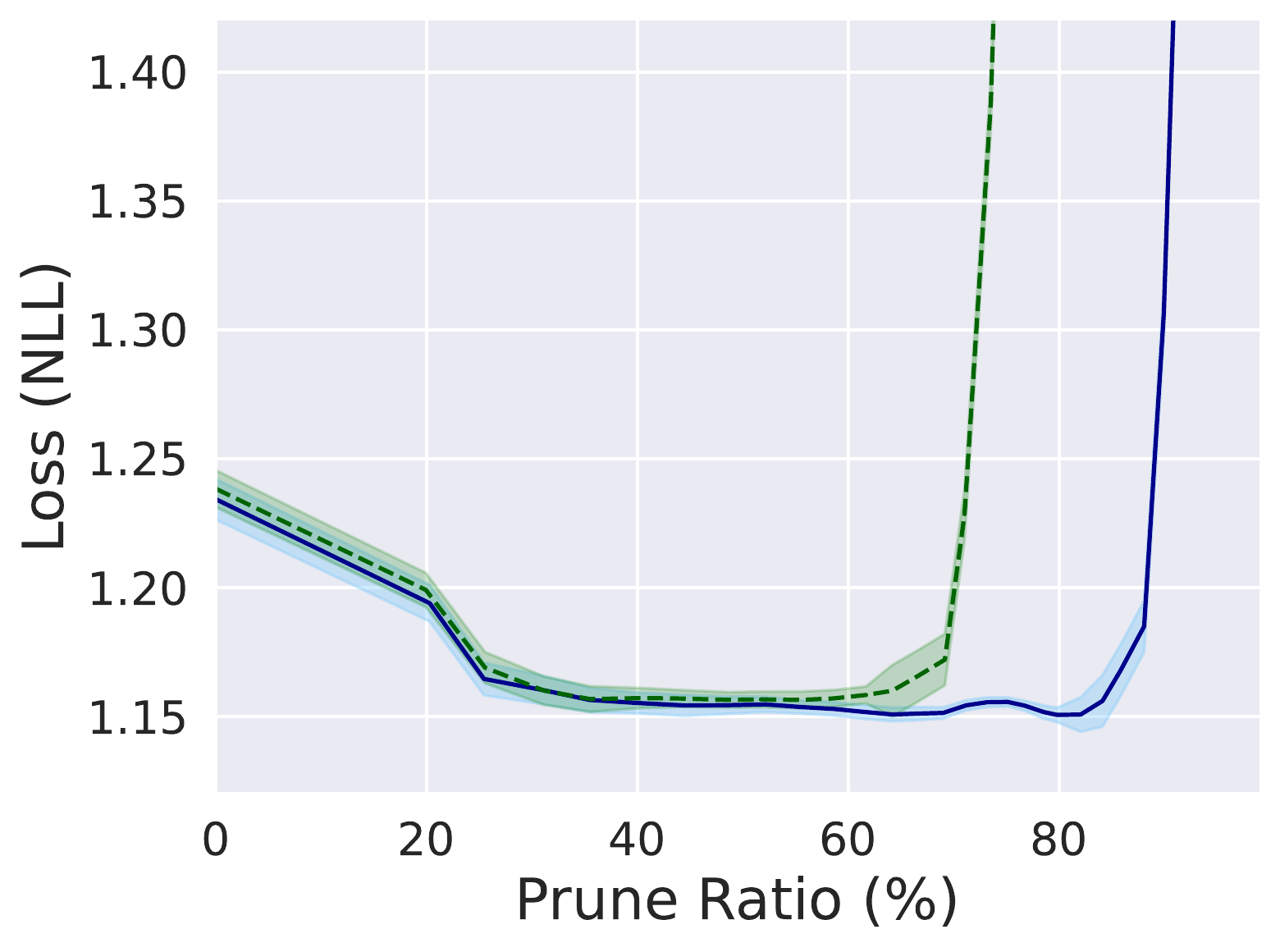}
    \subcaption{Gaussians}
\end{minipage}%
\begin{minipage}[t]{0.33\textwidth}
    \includegraphics[width=\textwidth]{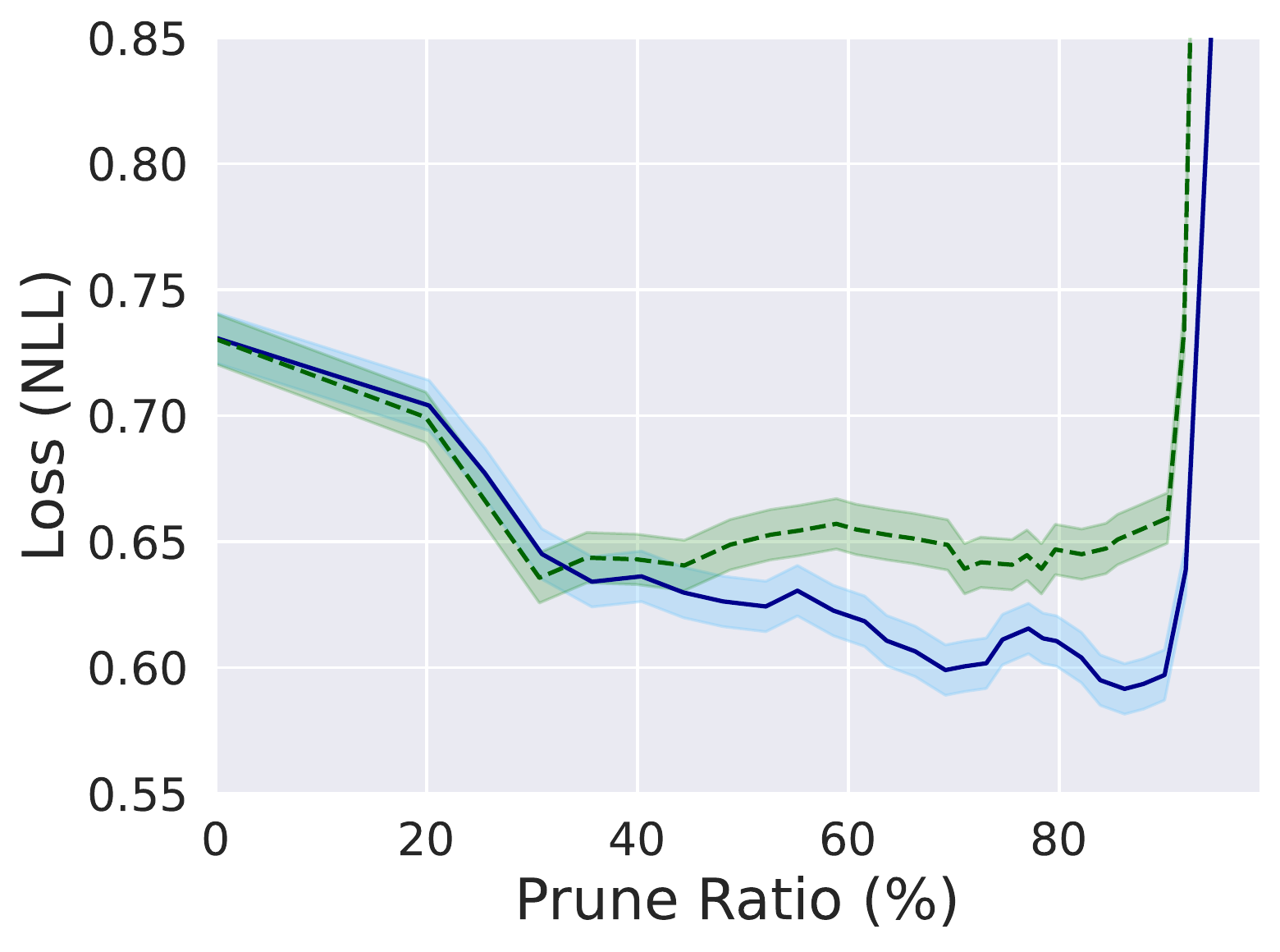}
    \subcaption{Gaussian Spiral}
\end{minipage}%
\begin{minipage}[t]{0.33\textwidth}
    \includegraphics[width=\textwidth]{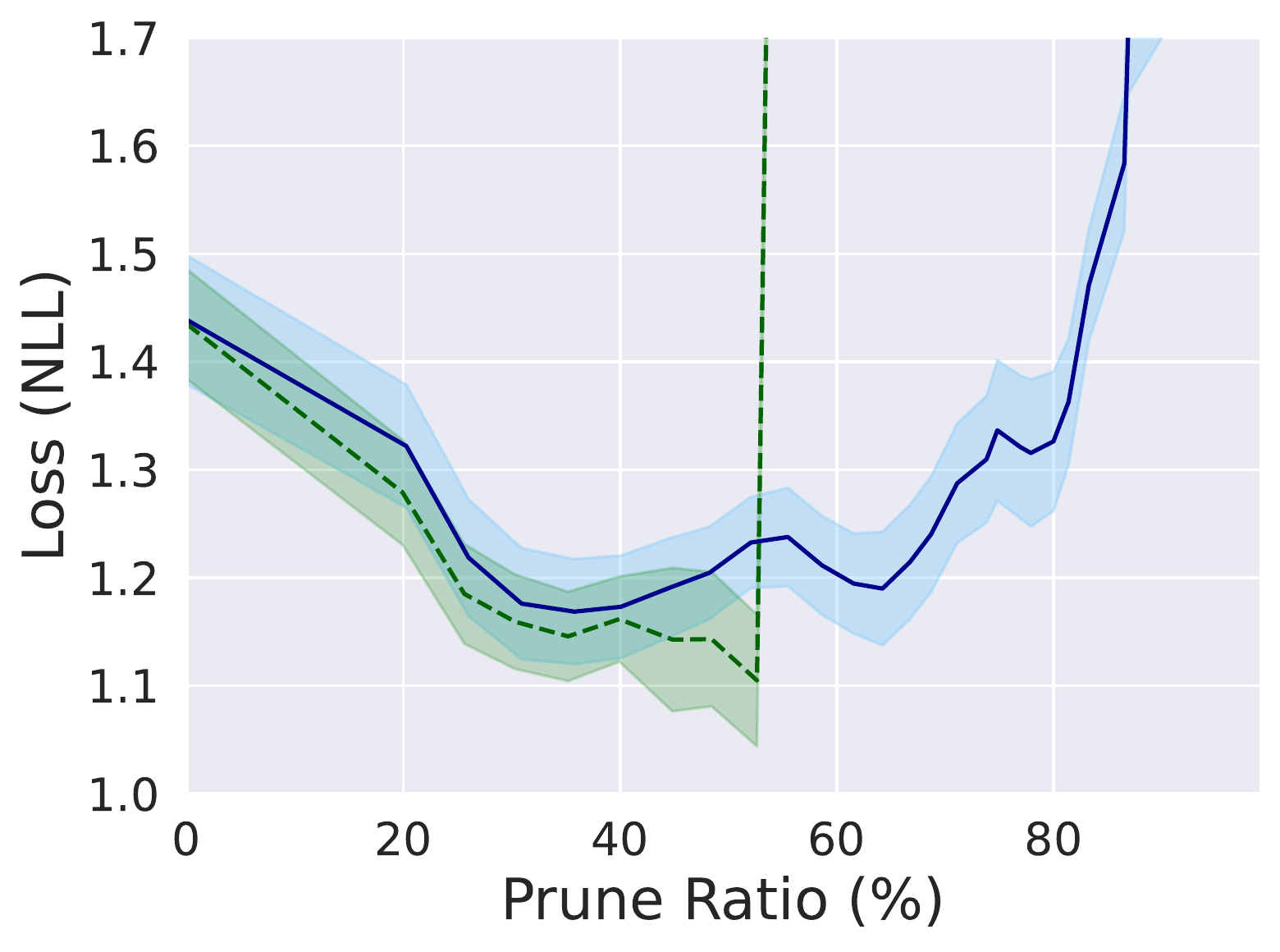}
    \subcaption{Spiral}
\end{minipage}%
\caption{Negative log likelihood of Sparse Flow as function of prune ratio. Sparse Flows were trained using regular CNFs instead of the FFJORD approximation. The remaining experimental setup follows Section~\ref{sec:experiments_2d}.
}
\label{fig-supp:toy_cnf_nll}
\end{figure}

Figure \ref{fig-supp:toy_cnf_nll} illustrates that densely connected flows (prune ratio = 0\%) might get stuck in sharp local minima and thus exhibit unfavorable generalization performance in terms of the NLL. Once we perform pruning, we observe that the quality of the density estimation in all tasks considerably improves. If we continue sparsifying the flows, depending on the task at hand, the flows get disrupted again.
Notably, we find that the straightforward Hessian computation may lead to slight improvements compared to using the FFJORD approximation, cf. results for the Spiral dataset using FFJORD and regular CNF as shown in Figure~\ref{fig:toy_nll}(c) and~\ref{fig-supp:toy_cnf_nll}(c), respectively. This may be expected in certain cases as  FFJORD replaces the exact Hessian computation with an approximation, thus potentially destabilizing training.

Overall, we can experimentally validate that our observations hold regardless of the specific method to compute the Hessian.

\section{Tabularized Results for Density Estimation on Real Data - Tabular}

\begin{table}[H]
\centering
\caption{Negative test log-likelihood (NLL) in nats on \textbf{\textsc{Power}} tabular dataset and corresponding architecture size in number of parameters and prune ratio for Sparse Flow (based on FFJORD). Results for unstructured and structured pruning are reported.}
\label{tab-supp:power}
\begin{tabular}{c|c|cc}
    \toprule
    \multicolumn{2}{c|}{\textbf{\textsc{Power}}} & \multicolumn{2}{c}{\textbf{Loss (nats)}}\\
    \midrule
    Prune ratio (\%) &	Number of parameters & Unstructured & Structured\\
    \midrule
    0\% & 43.3K & -0.34 &  -0.34 \\
    30\% & 30.3K & -0.48 &  -0.41 \\
    47\% & 23.0K & -0.50 &  -0.48 \\
    60\% & 17.4K & -0.51 &  -0.45 \\
    70\% & 13.2K & -0.55 &  -0.45 \\
    77\% & 9.95K & -0.55 &  -0.44 \\
    82\% & 7.65K & -0.52 &  -0.39 \\
    87\% & 5.81K & -0.45 &  -0.25 \\
    90\% & 4.43K & -0.39 & 0.04 \\
    \bottomrule
\end{tabular}
\end{table}




\begin{table}[H]
\centering
\caption{Negative test log-likelihood (NLL) in nats on \textbf{\textsc{Gas}} tabular dataset and corresponding architecture size in number of parameters and prune ratio for Sparse Flow (based on FFJORD). Results for unstructured and structured pruning are reported.}
\label{tab-supp:gas}
\begin{tabular}{c|c|cc}
    \toprule
    \multicolumn{2}{c|}{\textbf{\textsc{Gas}}} & \multicolumn{2}{c}{\textbf{Loss (nats)}}\\
    \midrule
    Prune ratio (\%) &	Number of parameters & Unstructured & Structured\\
    \midrule
    0\% & 279K & -8.64 & -8.64 \\
    30\% & 194K & -10.85 & -10.63 \\
    47\% & 147K & -11.15 & -10.93 \\
    60\% & 112K & -11.39 & -10.69 \\
    70\% & 84.6K & -11.59 & -10.20 \\
    77\% & 64.3K & -11.47 & -9.95 \\
    83\% & 48.6K & -10.85 & -9.85 \\
    87\% & 36.9K & -10.73 & -9.16 \\
    90\% & 28.0K & -10.03 & -7.58 \\
    \bottomrule
\end{tabular}
\end{table}

\begin{table}[H]
\centering
\caption{Negative test log-likelihood (NLL) in nats on \textbf{\textsc{Hepmass}} tabular dataset and corresponding architecture size in number of parameters and prune ratio for Sparse Flow (based on FFJORD). Results for unstructured and structured pruning are reported.}
\label{tab-supp:hepmass}
\begin{tabular}{c|c|cc}
    \toprule
    \multicolumn{2}{c|}{\textbf{\textsc{Hepmass}}} & \multicolumn{2}{c}{\textbf{Loss (nats)}}\\
    \midrule
    Prune ratio (\%) &	Number of parameters & Unstructured & Structured\\
    \midrule
    0\% & 547K & 17.54 & 17.54 \\
    20\% & 437K & 16.90 & 16.91 \\
    38\% & 340K & 16.56 & 16.54 \\
    52\% & 264K & 16.22 & 16.39 \\
    63\% & 205K & 16.00 & 16.21 \\
    71\% & 160K & 15.80 & 16.48 \\
    77\% & 124K & 15.67 & 16.48 \\
    82\% & 96.7K & 15.58 & 16.35 \\
    86\% & 75.5K & 15.59 & 16.97 \\
    89\% & 59.0K & 15.62 & 16.92 \\
    92\% & 46.4K & 15.99 & 17.34 \\
    93\% & 36.3K & 15.90 & 17.80 \\
    95\% & 28.9K & 16.04 & 18.77 \\
    \bottomrule
\end{tabular}
\end{table}

\begin{table}[H]
\centering
\caption{Negative test log-likelihood (NLL) in nats on \textbf{\textsc{Miniboone}} tabular dataset and corresponding architecture size in number of parameters and prune ratio for Sparse Flow (based on FFJORD). Results for unstructured and structured pruning are reported.}
\label{tab-supp:miniboone}
\begin{tabular}{c|c|cc}
    \toprule
    \multicolumn{2}{c|}{\textbf{\textsc{Miniboone}}} & \multicolumn{2}{c}{\textbf{Loss (nats)}}\\
    \midrule
    Prune ratio (\%) &	Number of parameters & Unstructured & Structured\\
    \midrule
    0\% & 821K & 10.38 & 10.38 \\
    20\% & 656K & 10.83 & 10.87 \\
    38\% & 510K & 11.11 & 10.93 \\
    52\% & 397K & 10.50 & 11.09 \\
    62\% & 308K & 10.77 & 11.10 \\
    71\% & 240K & 10.51 & 11.40 \\
    77\% & 186K & 10.64 & 11.07 \\
    82\% & 145K & 10.46 & 11.50 \\
    86\% & 112K & 10.37 & 11.35 \\
    89\% & 86.9K & 10.44 & 11.11 \\
    92\% & 67.7K & 10.60 & 10.93 \\
    94\% & 52.4K & 10.27 & 10.75 \\
    95\% & 40.8K & 10.05 & 10.41 \\
    96\% & 32.3K & 9.90 & 11.26 \\
    97\% & 24.4K & 10.15 & 12.09 \\
    98\% & 17.6K & 10.72 & 14.92 \\
    99\% & 9.13K & 13.61 & 39.01 \\
    \bottomrule
\end{tabular}
\end{table}




\begin{table}[H]
\centering
\caption{Negative test log-likelihood (NLL) in nats on \textbf{\textsc{Bsds300}} tabular dataset and corresponding architecture size in number of parameters and prune ratio for Sparse Flow (based on FFJORD). Results for unstructured and structured pruning are reported.}
\label{tab-supp:bsds300}
\begin{tabular}{c|c|cc}
    \toprule
    \multicolumn{2}{c|}{\textbf{\textsc{Power}}} & \multicolumn{2}{c}{\textbf{Loss (nats)}}\\
    \midrule
    Prune ratio (\%) &	Number of parameters & Unstructured & Structured\\
    \midrule
    0\% & 6.70M & -128.32 & -128.32 \\
    30\% & 4.69M & -145.54 & -145.42 \\
    47\% & 3.55M & -148.78 & -148.70 \\
    60\% & 2.68M & -149.96 & -149.92 \\
    70\% & 2.03M & -150.28 & -150.66 \\
    77\% & 1.54M & -151.11 & -150.11 \\
    83\% & 1.16M & -151.22 & -149.41 \\
    87\% & 878K & -151.13 & -149.42 \\
    90\% & 662K & -150.53 & -148.59 \\
    \bottomrule
\end{tabular}
\end{table}


\section{Tabularized Results for Density Estimation on Real Data - Vision}

Below, we report the tabularized results on MNIST and CIFAR10 for Sparse Flow using unstructured and structured pruning. Note that highly sparse networks can fail to converge beyond a certain prune ratio. Moreover, structured pruning may not converge for lower prune ratios compared to unstructured pruning since we prune away entire channels and neurons in the former.

\begin{table}[H]
\centering
\caption{Negative test log-likelihood (NLL) in bits/dim on \textbf{\textsc{MNIST}} and corresponding architecture size in number of parameters and prune ratio for Sparse Flow (based on FFJORD). Results for unstructured and structured pruning are reported.}
\label{tab-supp:mnist}
\begin{tabular}{c|c|cc}
    \toprule
    \multicolumn{2}{c|}{\textbf{\textsc{MNIST}}} & \multicolumn{2}{c}{\textbf{Loss (bits/dim)}}\\
    \midrule
    Prune ratio (\%) &	Number of parameters & Unstructured & Structured\\
    \midrule
0\% & 801K & 1.01 & 1.01 \\
20\% & 641K & 0.97 & 0.98 \\
38\% & 499K & 0.96 & 0.97 \\
52\% & 387K & 0.95 & 0.97 \\
62\% & 302K & 0.95 & 0.97 \\
71\% & 234K & 0.96 & 0.98 \\
77\% & 182K & 0.97 & 0.98 \\
82\% & 141K & 0.98 & 0.99 \\
86\% & 109K & 0.97 & 0.99 \\
89\% & 84.5K & 0.98 & 1.00 \\
    \bottomrule
\end{tabular}
\end{table}




\begin{table}[H]
\centering
\caption{Negative test log-likelihood (NLL) in bits/dim on \textbf{\textsc{CIFAR10}} and corresponding architecture size in number of parameters and prune ratio for Sparse Flow (based on FFJORD). Results for unstructured and structured pruning are reported. ``N/A'' indicates that Sparse Flow did not converge for the given prune ratio.}
\label{tab-supp:cifar10}
\begin{tabular}{c|c|cc}
    \toprule
    \multicolumn{2}{c|}{\textbf{\textsc{CIFAR10}}} & \multicolumn{2}{c}{\textbf{Loss (bits/dim)}}\\
    \midrule
    Prune ratio (\%) &	Number of parameters & Unstructured & Structured\\
    \midrule
    0\% & 1.36M & 3.45 & 3.45 \\
    20\% & 1.09M & 3.38 & 3.39 \\
    38\% & 845K & 3.37 & 3.38 \\
    52\% & 657K & 3.36 & 3.39 \\
    63\% & 510K & 3.37 & 3.40 \\
    71\% & 395K & 3.38 & N/A \\
    77\% & 308K & 3.39 & N/A \\
    82\% & 239K & 3.40 & N/A \\
    86\% & 186K & 3.42 & N/A \\
    89\% & 144K & 3.43 & N/A \\
    92\% & 112K & 3.45 & N/A \\
    94\% & 86.7K & 3.48 & N/A \\
    95\% & 67.4K & 3.50 & N/A \\
    \bottomrule
\end{tabular}
\end{table}


\section{More Hessian Analysis}
We performed the following additional Hessian experiments. We observe that our conclusions on the behavior of the Hessian is generalizable to other datasets as well.

\begin{table}[H]
    \centering
    \caption{Gaussians - Hessian Analysis (Structured Pruning)}
	\begin{tabular}{lcccc}
	\toprule
	Model & NLL & $\lambda_{max}(H)$ & $\text{tr}(H)$ & $\kappa(H)$ \\
	\midrule
Unpruned FFJORD	& 1.173 &	0.0190 &	0.098 &	48.2k \\
Sparse Flows(PR=25\%) &	1.157 &	0.0110 &	0.076 &	2.76k \\
Sparse Flows(PR=67\%) &	1.148 &	0.0090 &	0.560 &	15.17k \\
Sparse Flows(PR=82\%) &	1.120 &	0.0065 &	0.058 &	22.75k \\
Sparse Flows(PR=90\%) &	1.136 &	0.0035 &	0.033 &	4.70k \\
Sparse Flows(PR=94\%) &	1.173 &	0.0069 &	0.033 &	3.94k \\
Sparse Flows(PR=96\%) &	1.244 &	0.0071 &	0.043 &	0.58k \\
\bottomrule
    \end{tabular}
    \label{tab:gaussian_hessian}
\end{table}

\begin{table}[H]
    \centering
    \caption{Gaussians-Spiral - Hessian Analysis (Structured Pruning)}
	\begin{tabular}{lcccc}
	\toprule
	Model & NLL & $\lambda_{max}(H)$ & $\text{tr}(H)$ & $\kappa(H)$ \\
	\midrule	
Unpruned FFJORD	& 0.880 & 0.0130& 0.121 & 0.34k \\
Sparse Flows(PR=25\%) &	0.692 & 0.0076 & 0.058 & 0.76k \\
Sparse Flows(PR=48\%) &	0.634 &	0.0049 & 0.047 & 0.22k \\
Sparse Flows(PR=67\%) &	0.646 &	0.0052 & 0.051 & 0.75k \\
Sparse Flows(PR=82\%) &	0.657 &	0.0053 & 0.053 & 1.69k \\
Sparse Flows(PR=94\%) &	0.740 &	0.0086 & 0.070 & 0.11k \\
Sparse Flows(PR=96\%) &	0.986 &	0.0100 & 0.095 & 0.23k \\
\bottomrule
    \end{tabular}
    \label{tab:gaussian_spiral_hessian}
\end{table}

\section{Ablation Study}
In our ablation we study different types of features of neural ODE models.
We have prepared our systematic ablation study with results provided as part of the supplementary material in the proceedings available here: \url{https://openreview.net/forum?id=_1VZo_-aUiT}.

Please find the detailed description of this ablation study in the \verb|README.txt| file. We include this file's description here as well.

\subsection*{Overview}

We test different configurations by applying Sparse Flows (Algorithm~\ref{alg:prune}) to investigate what type of network configurations are most stable and robust with respect to pruning. For each type of sweep (ablation), we highlight one key study and one key result.

\subsection*{Sweep over Optimization Parameters}

\paragraph{Setup.} 
We study the stability of different configurations for the optimizer and how the different configurations affect the generalization performance during pruning.

\paragraph{Key observation.}
We can find the most stable parameter configuration for the optimizer by considering sparsifying the flow and thus inducing additional regularization. The most stable optimizer configuration is the one for which we can achieve the most pruning.

\subsection*{Sweep over Model Sizes - Depth vs. Width}

\paragraph{Setup.}
We study different network configurations with (approximately) the same number of parameters. The networks differ in the depth vs. width configuration. We test deep and narrow vs. shallow and wide.

\paragraph{Key observation.} 
Increasing the depth of the network while reducing the width of the network, in general, does not help improve the generalization performance of the network over different prune ratios. Specifically, one should pick the minimal depth of the RHS that ensures convergence. Usually, any depth beyond that does not help improve the generalization performance of the flow.

\subsection*{Sweep over Activations}

\paragraph{Setup.}
We study the same network configurations for the same amount of pruning and vary the activation function of the neural network on the RHS. As we prune, we hope to unearth which activation function is most robust to pruning and consequently to changes in the architecture.

\paragraph{Key observation.} 
ReLU is usually not a very useful activation function. Rather, some Lipschitz continuous activation functions are most useful. Generally, we found tanh and sigmoid to be most useful, although sigmoid was probably the most robust single configuration across all experiments

\subsection*{Sweep over ODE Solvers}

\paragraph{Setup.}
We study the same network configurations for the same amount of pruning and vary the ODE solver of the neural ODE flow. As we prune, we hope to unearth which solver is most robust to pruning and consequently to changes in the architecture.

\paragraph{Key observation.} 
Generally, we found adaptive step size solvers (dopri5) superior to fixed step size solvers (rk4, Euler). Moreover, we found backpropagation through time (BPTT) to be slightly more stable than the adjoint method. Interestingly enough, we could oftentimes only observe the differences between the robustness of the different solvers after we start pruning and sparsifying the flows.

\section{More on Pruning Configurations}
\textit{Iterative learning rate rewinding}: We use learning rate rewinding (LRR) which is a hyperparameter schedule for training/pruning/retraining of neural nets \citep{renda2020comparing} (see Algorithm \ref{alg:prune}).

\textit{Pre-defined pruning threshold:} We pick a desired prune ratio and prune the weights with the smallest magnitudes until we obtain it. The largest weight that is being pruned constitutes the pre-defined threshold for pruning.

\textit{What kinds of structures are being considered in structured pruning?}  We consider neurons in fully-connected layers and channels with their corresponding filters in convolutional layers for structured pruning. The corresponding pruning score is the norm of the neuron/channel weights as specified in Table 1.

\section{Reproducibility Matters}
All code and data which contains the details of the hyperparameters used in all experiments are openly accessible online at: \url{https://github.com/lucaslie/torchprune} For the experiments on the toy datasets, we based our code on the TorchDyn library~\citep{poli2020torchdyn}. For the experiments on the tabular datasets and image experiments, we based our code on the official code repository of FFJORD~\citep{grathwohl2019ffjord}.

\end{document}